\documentclass[10pt,twocolumn,letterpaper]{article}

\usepackage{iccv}
\usepackage{times}
\usepackage{epsfig}
\usepackage{graphicx}
\usepackage{amsmath}
\usepackage{amssymb}
\usepackage{color}
\usepackage{multirow}
\usepackage{booktabs}
\usepackage{bbm}
\usepackage{bbding}
\usepackage{makecell}
\usepackage{colortbl}
\usepackage{overpic}
\usepackage{caption}
\usepackage{subcaption}
\usepackage{bm}
\usepackage{algorithm}
\usepackage{algpseudocode}
\usepackage{stfloats}

\definecolor{othercolor}{rgb}{0.8 , 0.4 , 0.9}
\definecolor{RowColor}{RGB}{219,242,230}
\definecolor{ggreen}{RGB}{0,165,79}
\definecolor{TableHead}{RGB}{214,226,239}
\definecolor{yyellow}{rgb}{1 , 0.7 , 0}

\def\fullamount{\textcolor{yyellow}{$\spadesuit$}}

\usepackage[pagebackref=true,breaklinks=true,letterpaper=true,colorlinks,bookmarks=false]{hyperref}

\iccvfinalcopy 

\def\iccvPaperID{****} 

\ificcvfinal\fi

\begin{document}

\title{Deficiency-Aware Masked Transformer for Video Inpainting}

\author{%
    Yongsheng Yu$^{1}$ \hspace{12pt} Heng Fan$^{2}$ \hspace{12pt} Libo Zhang$^{1}$$^\dagger$\\
    $^{1}$Institute of Software, Chinese Academy of Sciences\\
    $^{2}$Department of Computer Science and Engineering, University of North Texas\\
    \texttt{yonsheng.yu@gmail.com; heng.fan@unt.edu; libo@iscas.ac.cn}\\
}

\twocolumn[{%
\renewcommand\twocolumn[2][]{#1}%
\maketitle%
\vspace{-0.4cm}

\centering \centering
\includegraphics[width=\textwidth]{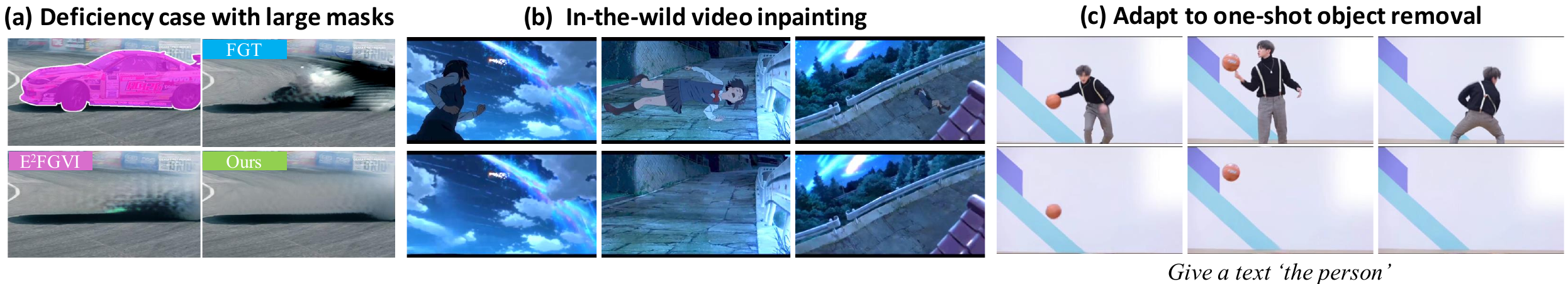}
\captionsetup[figure]{aboveskip=0.2cm}
\captionof{figure}
{
(a) Video inpainting result and a comparison with state-of-the-art models for handling large mask. 
(b) Generalizing to anime video clips that is outside the training dataset.
(c) The proposed model easily adapts to one-shot object removal by accepting text or stroke input only.
\emph{Best viewed in color and by zooming in for all figures throughout the paper.
}
}
\vspace*{0.2cm}
\label{fig:teaser}
\vspace{0.25cm}
}] 


\ificcvfinal\thispagestyle{empty}\fi

\begin{abstract}
\vspace{-0.1cm}

Recent video inpainting methods have made remarkable progress by utilizing explicit guidance, such as optical flow, to propagate cross-frame pixels. However, there are cases where cross-frame recurrence of the masked video is not available, resulting in a deficiency. In such situation, instead of borrowing pixels from other frames, the focus of the model shifts towards addressing the inverse problem.
In this paper, we introduce a dual-modality-compatible inpainting framework called Deficiency-aware Masked Transformer (DMT), which offers three key advantages. Firstly, we pretrain a image inpainting model DMT$_{\text{img}}$ serve as a prior for distilling the video model DMT$_{\text{vid}}$, thereby benefiting the hallucination of deficiency cases. Secondly, the self-attention module selectively incorporates spatiotemporal tokens to accelerate inference and remove noise signals. Thirdly, a simple yet effective Receptive Field Contextualizer is integrated into DMT, further improving performance.
Extensive experiments conducted on YouTube-VOS and DAVIS datasets demonstrate that DMT$_{\text{vid}}$ significantly outperforms previous solutions. The code and video demonstrations can be found at \url{github.com/yeates/DMT}.

\end{abstract}

\section{Introduction}

Video inpainting, aiming to reconstruct the corrupted regions with coherent and consistent contents in videos, is an important problem in computer vision and has many applications, including object removal~\cite{kim19deep,chang19free,li22e2fgvi}, video restoration~\cite{lee19copy}, etc. Despite considerable progress in recent years, accurate video inpainting remains an open problem because of the complex and challenging scenarios in videos.

One major challenge in video inpainting is the deficiency case~\cite{ouyang21iivi}, where the masked content is absent throughout the whole video. 
In such a case, the inpainting models often degrade into an inverse problem. This issue becomes more severe when dealing with large mask, as existing video inpainting models may fail and produce artifacts (see Fig.~\ref{fig:teaser}(a)). In contrast, generative image inpainting~\cite{zhao21comod,rombach22ldm,yu22invertfill,jain22fcf} offers a high-fidelity solution to address the deficiency case by leveraging the context of unmasked image regions. Given the close interrelation between image and video inpainting, a natural question arises: \emph{Can we leverage image inpainting techniques to enhance video inpainting in deficiency cases?} In simpler terms, \emph{Can we pre-train image inpainting models to improve video inpainting?}

We answer \emph{yes}. However, it is \emph{non-trivial}. The domain gap between video and image inpainting tasks arises from their different objectives. Video models attempt to borrow accurate pixel information across frames, whereas image models are learnt to fill holes based on the unmasked regions. Additionally, extending a 2D module to handle video inpainting in a simplistic manner, such as utilizing 3D convolutions~\cite{chang19free,wang19video} and vanilla self-attention~\cite{zeng20sttn,liu21fuseformer} in the generator, lacks task-specific exploration.

\begin{figure}[!t]
\centering
\includegraphics[width=\linewidth]{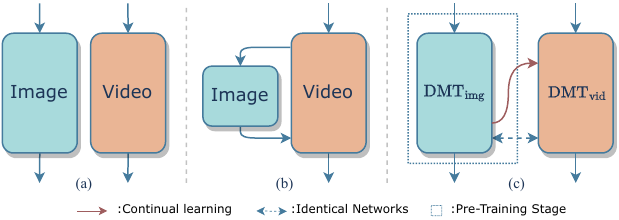}
\caption{Illustration of our idea. Image (a) shows video and image models are independent of each other, which is the most common way currently; image (b) shows an image inpainting model is used for video inpainting task~\cite{xu19dfc}. Image (c) shows the bridging of image and video inpainting by an efficient pre-training and continual learning strategy.}
\label{fig:rela}
\vspace{-2mm}
\end{figure}

\noindent
\textbf{Our solution.} In order to conquer this barrier and wipe out the design and domain gap between the image and the video inpainting, we introduce a novel dual-modality-compatible inpainting framework that is flawlessly compatible with image and video inputs. To leverage the hallucination ability of image inpainting model, we empirically use pretrained DMT$_{\text{img}}$ as a prior to preserves knowledge in dealing with deficiency problems. Additionally, we train DMT$_{\text{vid}}$ from scratch and apply a continual learning loss, which transfers knowledge from the DMT$_{\text{img}}$ prior and trades off the plasticity and stability~\cite{mermillod2013stability} when learning from new video data. Fig.~\ref{fig:rela} shows our idea. 

At the core of our framework is the Deficiency-aware Masked Transformer, which models long-range dependencies across the spatial or temporal axis for images sequences. Unlike the vanilla transformer, DMT incorporates a Token Selection mechanism to drop invalid tokens where all internal pixels fall within the mask region. To iteratively activate invalid tokens, we introduce a heuristic Mask Activation algorithm for self-attention and convolution operators. Furthermore, to benefit from a high receptive field for inpainting, we integrate a simple yet effective network called Receptive Field Contextualizer (RFC) into DMT.

Our approach offers several advantages compared to previous video inpainting models. First, it seamlessly inherits the hallucination ability from the pre-trained image inpainting model, DMT$_{\text{img}}$, enabling it to generate content for deficienct region. Second, the Token Selection mechanism and Mask Activation algorithm reduce computational cost, as the complexity of the transformer is proportional to the input mask ratio, while slightly improving performance by removing noise tokens. Third, the RFC expands the receptive field of DMT and learns high-frequency signals~\cite{park22how}, combining the strengths of both transformers and convolutional networks for better results.

To validate our proposed framework, we conduct extensive experiments on YouTube-VOS~\cite{youtubevos} and DAVIS~\cite{davis}. The results demonstrate that our method outperforms state-of-the-art video inpainting approaches, setting new records. Specifically, compared to the SOTA method~\cite{li22e2fgvi} in terms of PSNR, our method achieves an increase of 0.81 dB on DAVIS and 0.56 dB on YouTube-VOS, respectively. Moreover, our model generalizes well to in-the-wild video inpainting scenarios and can be easily adapted to a one-shot object removal pipeline without frame-wise masks input (see Fig.\ref{fig:teaser}(b) and (c)).

\section{Related works}
\noindent
\textbf{Image Inpainting.} Classical image inpainting~\cite{barnes09patchmatch} employs heuristic methods to search for and propagate information from reference regions to fill in missing pixels. Following the seminal inpainting work~\cite{pathak16ce} using deep learning, many extensions have been introduced for improvements, such as multi-step sampling~\cite{song21scoresde,yu2023magic}, auxiliary prior guidance~\cite{yu22mmt,yu2023magic}, generator-discriminators~\cite{jain22fcf,yu22invertfill}, mask-aware network designs~\cite{Iizuka17global,Liu18partial,yu19free,yu20reigon}, etc. More recently, inspired by the power of Transformer~\cite{vaswani2017attention,Dosovitskiy21ViT}, it has been leveraged for inpainting by modeling the long-range dependencies in images and exhibited promising performance~\cite{li22mat,yu22mmt,zeng20sttn}. Despite this, the vanilla Transformer might suffer from high computational complexity, limiting its efficiency. Similar to these approaches, we also utilize Transformer for capturing long-range dependency. But \emph{differently}, our framework employs a Token Selection mechanism and Mask Activation algorithm to drop and activate invalid tokens, improving efficiency and boosting performance. 

\begin{figure*}[t]
  \centering
  \includegraphics[width=0.95\textwidth]{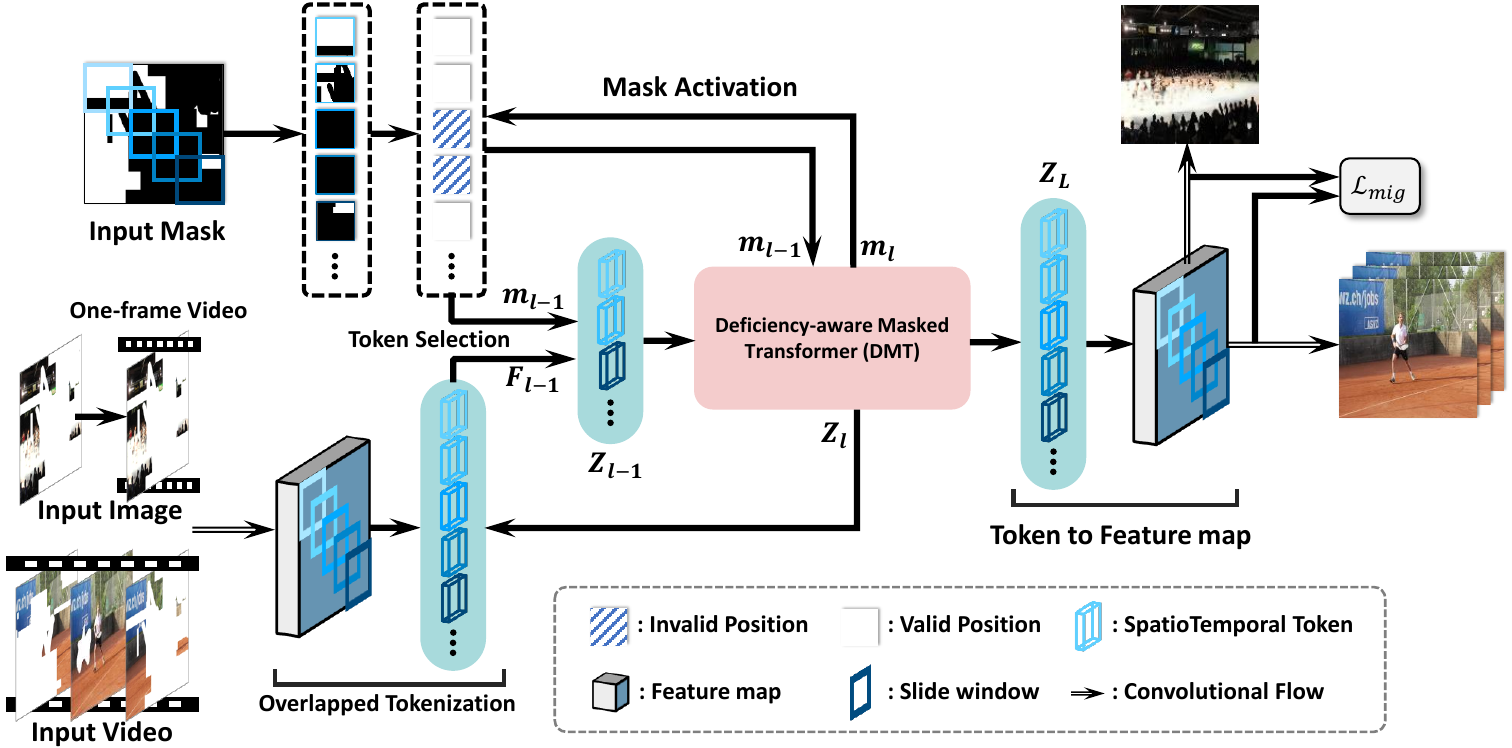}
  \caption{Our proposed inpainting framework consists of the following components: DMT bottlenecks, token selection operator, mask activation strategy, migration regularization, and 2D-convolutional encoder and decoder (shown as convolutional flow in the figure). In each successive DMT layer, masks are dynamically hallucinated until they vanish. }\label{fig:pipeline}
  \vspace{-3mm}
\end{figure*}

\vspace{0.5em}
\noindent
\textbf{Video Inpainting.} Benefiting from deep learning, video inpainting~\cite{kim19deep} has witnessed great progress in recent years. Compared with image inpainting, video inpainting needs to deal with the additional time dimension, which introduces challenges such as camera movement, temporal consistency preserving, and reference between frames. As a result, it is hard to apply image inpainting models for video inpainting in generating temporally coherent content. In order to adapt the video domain, great efforts have been devoted to designing 3D convolutions~\cite{chang19free,wang19video}, optical flow~\cite{xu19dfc,zhang22inertia,kang22error,zhang22flow} and temporal transformers~\cite{zeng20sttn,liu21fuseformer,zhang22flow,li22e2fgvi,cai22devit}, showing excellent performance. Among these works, E$^2$FGVI~\cite{li22e2fgvi}, FGT~\cite{zhang22flow}, and DeViT~\cite{cai22devit} are the most recent leading models. 
E$^2$FGVI~\cite{li22e2fgvi} proposes end-to-end flow-guided feature propagation to enhance its temporal focal transformer~\cite{yang21focal}. FGT~\cite{zhang22flow} integrates the flow completion network into the transformer to decouple self-attention along temporal and spatial perspectives. DeViT~\cite{cai22devit} designs spatial and temporal branches of transformers with patch-wise alignment and matching in order to adapt various motion scenarios. \emph{Different} than these approaches, we introduce Receptive Field Contextualizer into DMT that enjoy strong strength of both transformer and CNN.

\vspace{0.5em}
\noindent
\textbf{Masked Visual Modeling.} Mask modeling~\cite{devlin19bert} comes from natural language processing. The work of Masked Autoencoders~\cite{he22mae} (MAEs) proposes to learn representations by recreating the original images from patch-form masked images. Several studies~\cite{DBLP:journals/corr/abs-2204-01678,DBLP:journals/corr/abs-2208-10442,DBLP:conf/iclr/Bao0PW22} observe that this self-supervised pre-training is beneficial for improving transformers. Our work shares similar spirits with these variants of MAEs that only encodes unmasked tokens for long-range dependency modeling. The benefit of employing a masked transformer is to lessen the computational load, especially for large masks. 

\vspace{0.5em}
\noindent
\textbf{Pre-training in Vision.} Pre-training is popular in computer vision to transfer knowledge between tasks. For example, the classification networks pre-trained on ImageNet~\cite{deng2009imagenet} are often employed for other various tasks such as object  detection~\cite{ren2015faster}, segmentation~\cite{long2015fully}, etc. In these tasks, the backbone of the pre-trained will be borrowed for initializing the model in the new task. Our method shares a similar idea but is \emph{different}. First, we train a new video model from scratch, instead of finetune pre-trained backbone. In addition, we introduce a continual learning loss to trade off the plasticity and stability~\cite{mermillod2013stability}. 

\section{The Proposed Methodology}
Given a masked sequence $\{X_t \in \mathbb{R}^{3 \times H\times W} | t \in [1,T]\}$ with sequence length $T$ and corresponding frame-wise binary masks $\{M_t \in \mathbb{R}^{1 \times H\times W} | t \in [1,T]\}$, we aim to hallucinate visually appealing and semantically plausible contents in space and time dimensions for missing regions. 
The task degenerates to image inpainting while the time step of the frame sequence is set to $1$. 
The proposed inpainting framework, which infers context features across spatial and temporal adaptively and reconstructs them to output $\{\hat{Y}_t \in \mathbb{R}^{3 \times h\times w} | t\in [1,T]\}$. Our approach works flawlessly with image and video inputs, and we ignore time step $t$ below for simplicity’s sake.

\subsection{Overall Architecture}

The pipeline of our proposed approach is illustrated in Figure~\ref{fig:pipeline}. It consists of a 2D-convolutional encoder-decoder and a Deficiency-aware Masked Transformer. The architecture operates as follows:
The encoder encodes the masked input, reducing its size by a factor of $4$ and producing $C$ channel convolutional feature maps $X_{\downarrow} \in \mathbb{R}^{C \times \frac{H}{4} \times \frac{W}{4}}$.
Next, we tokenize the $X_{\downarrow}$ feature map using a linear network that maps the feature dimension $C$ to $d$. Tokens from all frames are merged into the same dimension, resulting in spatiotemporal tokens represented by $F\in \mathbb{R}^{N \times d}$.
The Deficiency-aware Masked Transformer (DMT) acts as the bottleneck blocks with $L$ layers to generate missing content using valid-only spatiotemporal tokens $Z_l, l \in [0, L-1]$. This enables the learning of long-range dependencies across both temporal and spatial dimensions. Finally, we inverse-tokenize $Z_L$ to obtain the feature map and reconstruct completed video frames $\hat{Y}$ using the decoder.

{\noindent\bf Token Selection.} In previous transformer-based methods, all tokens are considered to have equal importance. However, tokens extracted from the missing region provide insignificant knowledge and impede computational speed.
We hence drop all tokens within the masked region and only input the valid tokens $Z_0$ into the DMT blocks. We define the Token Selection process as $\phi(\cdot, \cdot)$, which yields valid-only tokens:
\begin{equation}
Z = \phi(F, m), \quad N^\prime = |\phi(F, m)|,
\end{equation}
Here, $F \in \mathbb{R}^{N\times d}$ and $Z \in \mathbb{R}^{N^\prime\times d}$ represent the full-amount tokens and valid-only tokens, respectively. $N^\prime$ denotes the number of tokens after removing the invalid ones.
Since our method handles free-form masks rather than patch-form masks, we consider an image token as valid when its corresponding pixels are not fully masked. To indicate the masking status of each pixel explicitly, the downscaled mask $m \in \mathbb{R}^{1 \times \frac{H}{4} \times \frac{W}{4}}$ is sent to the DMT blocks along with tokenized sequence $Z$.
As a result, the DMT blocks learn valid-only tokens instead of full-amount tokens. This removal of noise signals from invalid tokens speeds up DMT.

\begin{figure}[t]
  \centering
  \includegraphics[width=0.75\columnwidth]{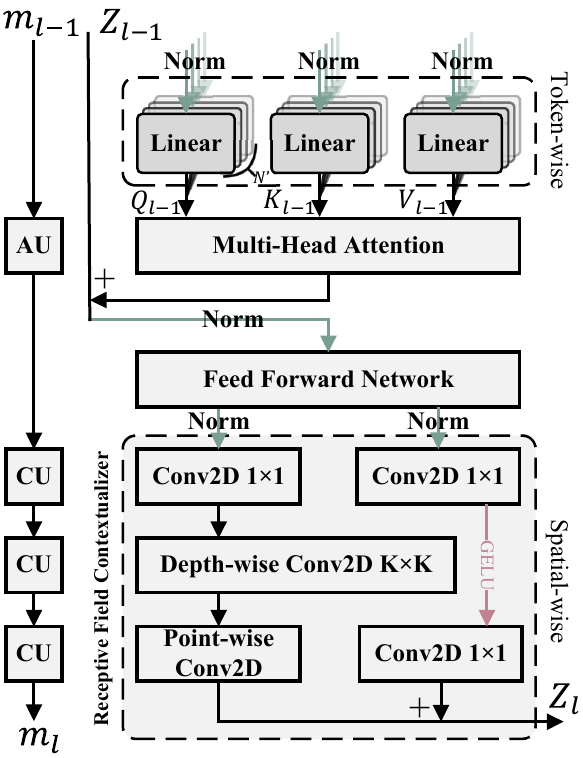}
  \caption{Illustration of Deficiency-aware Masked Transformer. Blocks `AU' and `CU' represent Attention mask Updater and Convolution mask Updater modules of mask activation, respectively.}\label{fig:stem}
  \vspace{-3mm}
\end{figure}

\subsection{Deficiency-aware Masked Transformer}\label{sec:dmt}
In this section, we introduce the Deficiency-aware Masked Transformer, which offers a more effective alternative to the commonly used vanilla vision transformer~\cite{Dosovitskiy21ViT,liu21fuseformer}. The DMT consists of a masked transformer that specifically handles valid tokens, as well as novel components such as Mask Activation strategy and Receptive Field Contextualizer module.

The Masked Transformer is borrowed from the vanilla transformer and consists of multi-head self-attention and a feed-forward network. Given the input valid-only tokens $Z_l$ at the $l$-th stack, where $l \in [1, L-1]$ and $L$ is the number of stacked DMT blocks, the front part of a DMT block can be formulated as follows:
\begin{subequations}
\begin{align}
    Z^\prime_{l-1} &= \text{MSA}(\text{LN}_1(Z_{l-1})) + Z_{l-1}, \\
    \overline{Z}_{l-1} &= \text{FFN}(\text{LN}_2(Z^\prime_{l-1})) + Z^\prime_{l-1},
\end{align}
\end{subequations}
Here, MSA and LN denote the standard multi-head self-attention~\cite{Dosovitskiy21ViT} and layer normalization~\cite{ba16layernorm}, respectively. We use FFN~\cite{Dosovitskiy21ViT} with token-feature alternately warping~\cite{yuan21t2t,liu21fuseformer} to establish connections between embedded tokens.

{\noindent\bf Mask Activation.} Learning with valid-only tokens can alleviate the computational strain of the vanilla transformer, especially when dealing with large-scale masks. However, the dropped tokens cannot be automatically reconstructed by the Masked Transformer.
Hence, we propose a heuristic Mask Activation strategy, which ensures the hallucination of all invalid tokens at the end of the DMT bottleneck. The Mask Activation explicitly changes the validity of invalid tokens based on a simple rule: self-attention and convolution operators will reconstruct masked pixels. Thus, we simulate these operators to activate the corresponding tokens.

Specifically, we introduce a collection of mask updaters, which are used for the convolution and self-attention operators, respectively. As summarized in Algorithm~\ref{algo:mask_activation}, the mask updater first tokenizes the current mask using a sliding window with the same parameters as the operator. Then, it re-normalizes each token $\mathrm{P}$ according to the following rule:
\begin{equation}\label{eqn:soft_min}
\mathrm{P}_{x,y}=1, \text{ iff sum}(\mathrm{P}) > 0.
\end{equation}

The tokenized mask $\mathrm{P}$ is binary-valued, similar to $m$, where $1$ represents an unmasked pixel at the location ($x$, $y$). Finally, the mask map validity is updated by rearranging the tokens into a feature map.
The mask updaters simulate the dynamics of hallucination, thereby iteratively activating invalid tokens in the DMT blocks.

\begin{algorithm}[b]
\small
\caption{Mask Activation}\label{algo:mask_activation}
 \noindent\textbf{Input:} $m\in \mathbb{R}^{C\times H\times W}$, $k$, $s$, $p$: mask map and sliding window parameters including kernel size, stride, and padding \\
 \noindent\textbf{Require:} intermediate token sequence $\overline{m} \in \mathbb{R}^{Ck^2 \times N}$, number of tokens $N = \lfloor \frac{H+2p-k}{s} \rfloor + 1 \times \lfloor \frac{W+2p-k}{s} \rfloor + 1$, and token index $n \in [0, N-1]$   \\
 \noindent\textbf{Output:} $\hat{m}\in \mathbb{R}^{C\times H\times W}$: updated mask map
\begin{algorithmic}[1]

\State {Initialize padded mask map $m \gets \text{\tt pad}(m, p)$, zero-filled token sequence $\overline{m} \gets \text{\tt zeros\_like}(Ck^2, N)$, and $\hat{m} \gets \text{\tt zeros\_like}(C, H+2p, W+2p)$}
\State \textbf{Align Mask Sequence:}
\For {$(n, i, j)$ in $\text{\tt enumerate}(\text{\tt range}(1, H+2p-k+1, s), \text{\tt range}(1, W+2p-k+1, s))$} 
  \State $\text{Token} \gets m[:,i:i+k,j:j+k]$ 
  \State $\overline{m}[:,n] \gets \text{\tt reshape}(\text{Token}, Ck^2)$
\EndFor
\State \textbf{Re-normalize Validity:}
\State $\overline{m} \gets \text{\tt ones\_like}(\overline{m}) \times (\text{\tt sum}(\overline{m}, \text{axis}=0) > 0)$ (Eqn. \ref{eqn:soft_min})
\State \textbf{Update Mask Map:}
\For {$(n, i, j)$ in $\text{\tt enumerate}(\text{\tt range}(1, H+2p-k+1, s), \text{\tt range}(1, W+2p-k+1, s))$} 
\State $\hat{m}[:,i:i+k,j:j+k] \mathrel{+}= \text{\tt reshape}(\overline{m}[:,n], C, k, k)$ 
\EndFor
\State $\hat{m} \gets \text{\tt clamp}(\text{\tt unpad}(\hat{m}, p), 0, 1)$ 
\end{algorithmic}
\end{algorithm}

{\noindent\bf Receptive Field Contextualizer.} 
Tokenizing feature maps into patches can result in the loss of high-frequency details and spatial structure in the image. To address this issue, we propose a simpler and stronger Receptive Field Contextualizer (RFC). The RFC module reconstructs spatial features, extracts high-frequency semantic details, and embeds learnable positions implicitly~\cite{wu2021cvt}.

As shown in Figure~\ref{fig:stem}, the RFC first employs skip connections to preserve temporal correlation and low-level feature information. It then utilizes two parallel branches to reconstruct spatial information. Within each branch, a $1\times 1$ convolution integrates features across channels. One branch applies Gaussian Error Linear Unit (GELU) to capture fine-grained local details at a smaller scale and enhance non-linear representation capability. The other branch utilizes depthwise separable convolution~\cite{chollet2017xception} with a large kernel size $K \times K$. This approach enhances the convolutional receptive field without excessively increasing the number of parameters. We integrate the RFC module into the Masked Transformer, and the rest of a DMT block is as follows:
\begin{equation}
Z_{l} = \text{RFC}(\overline{Z}_{l-1}) + \overline{Z}_{l-1}.
\end{equation}

The token feature is reformed into a spatial size because it was first flattened before being fed into the RFC. Detailed analysis and comparisons with similar high-receptive-field FFC modules~\cite{suvorov22lama} are provided in Table~\ref{tab:temporal_ii_sota} and~\ref{tab:abla_largeks}, highlighting the effectiveness of our approach.

\subsection{Migration Regularization}
To train a video inpainting model DMT$_{\text{vid}}$, we first pre-train a DMT$_{\text{img}}$ on the YouTube-VOS dataset~\cite{youtubevos}, which we consider as an image dataset. We penalize the image inpainting model DMT$_{\text{img}}$ using adversarial loss, perceptual loss, and $R_1$ regularization loss, following the approach in~\cite{li22mat}, which preserves the knowledge in coping with deficiency problems.

Building upon the pre-trained DMT$_{\text{img}}$, we proceed to train DMT$_{\text{vid}}$ on the video dataset YouTube-VOS~\cite{youtubevos} from scratch. We supervise the model DMT$_{\text{vid}}$ using various losses, such as the reconstruction loss (L1 distance) and the adversarial loss, following~\cite{li22e2fgvi}. To better exploit the knowledge from the pre-trained DMT$_{\text{img}}$, we introduce a Migration Regularization term $\mathcal{L}_{mig}$ to facilitate continual learning. This regularization term propagates context encoding features from the pre-trained model to the video model, and it is defined as:
\begin{equation}\label{eqn:mig}
\mathcal{L}_{mig}=
\sum_{l=1}^L \| m \odot \left( h^{(l)}_\theta - \text{ReLU} (\hat{h}^{(l)}) \right)\|_2^2,
\end{equation}

where $h_\theta^{(l)}$ and $\hat{h}^{(l)}$ indicates the $l$-th layer output features of video model and pre-trained model, respectively.

\begin{table}[b]
\small
\centering
\caption{Quantitative comparisons with SOTAs of video inpainting methods on free-form masks. 
$\uparrow$ indicates higher is better, and $\downarrow$ indicates lower is better.
The \textbf{best} and \underline{second best} results are in bold and underline.}
\label{tab:main_video}
\setlength{\tabcolsep}{3.8pt}
\begin{tabular}{l|cc}
\toprule \rowcolor{TableHead} 
 & \multicolumn{2}{c}{PSNR$\uparrow$/SSIM$\uparrow$/VFID$\downarrow$} \\ \rowcolor{TableHead}  
Method                        & \multicolumn{1}{c|}{DAVIS}       &     \multicolumn{1}{c}{Youtube-VOS}                                \\ \cmidrule(r){1-1} \cmidrule(r){2-2} \cmidrule(r){3-3}
VINet~\cite{kim19deep}  & 28.96 / 0.941 / 0.199 & 29.20 / 0.943 / 0.072   \\
DFVI~\cite{xu19dfc} & 28.81 / 0.940 / 0.187 & 29.16 / 0.943 / 0.066 \\
LGTSM~\cite{chang19lgstm} & 28.57 / 0.941 / 0.170 & 29.74 / 0.950 / 0.070 \\
CAP~\cite{lee19copy} & 30.28 / 0.952 / 0.182 & 31.58 / 0.961 / 0.071 \\
FGVC~\cite{gao20fgvc} & 30.80 / 0.950 / 0.165 & 29.67 / 0.940 / 0.064 \\
STTN~\cite{zeng20sttn} & 30.67 / 0.956 / 0.149 & 32.34 / 0.966 / 0.053 \\
FuseFormer~\cite{liu21fuseformer} & 32.54 / 0.970 / 0.138 & 33.29 / 0.968 / 0.053 \\
FGT~\cite{zhang22flow} & \underline{33.23} / 0.966 / 0.138 & 32.25 / 0.960 / 0.055 \\
E$^2$FGVI~\cite{li22e2fgvi} & 33.01 / \underline{0.972} / \underline{0.116} & \underline{33.71} / \underline{0.970} / \underline{0.046} \\ \cmidrule(r){1-1} \cmidrule(r){2-2} \cmidrule(r){3-3} 
\rowcolor{RowColor} Ours & \textbf{33.82} / \textbf{0.976} / \textbf{0.104} & \textbf{34.27} / \textbf{0.973} / \textbf{0.044} \\ \bottomrule
\end{tabular}
\end{table} 

\begin{figure*}[!t]
  \centering
    \begin{overpic}[width=\textwidth]{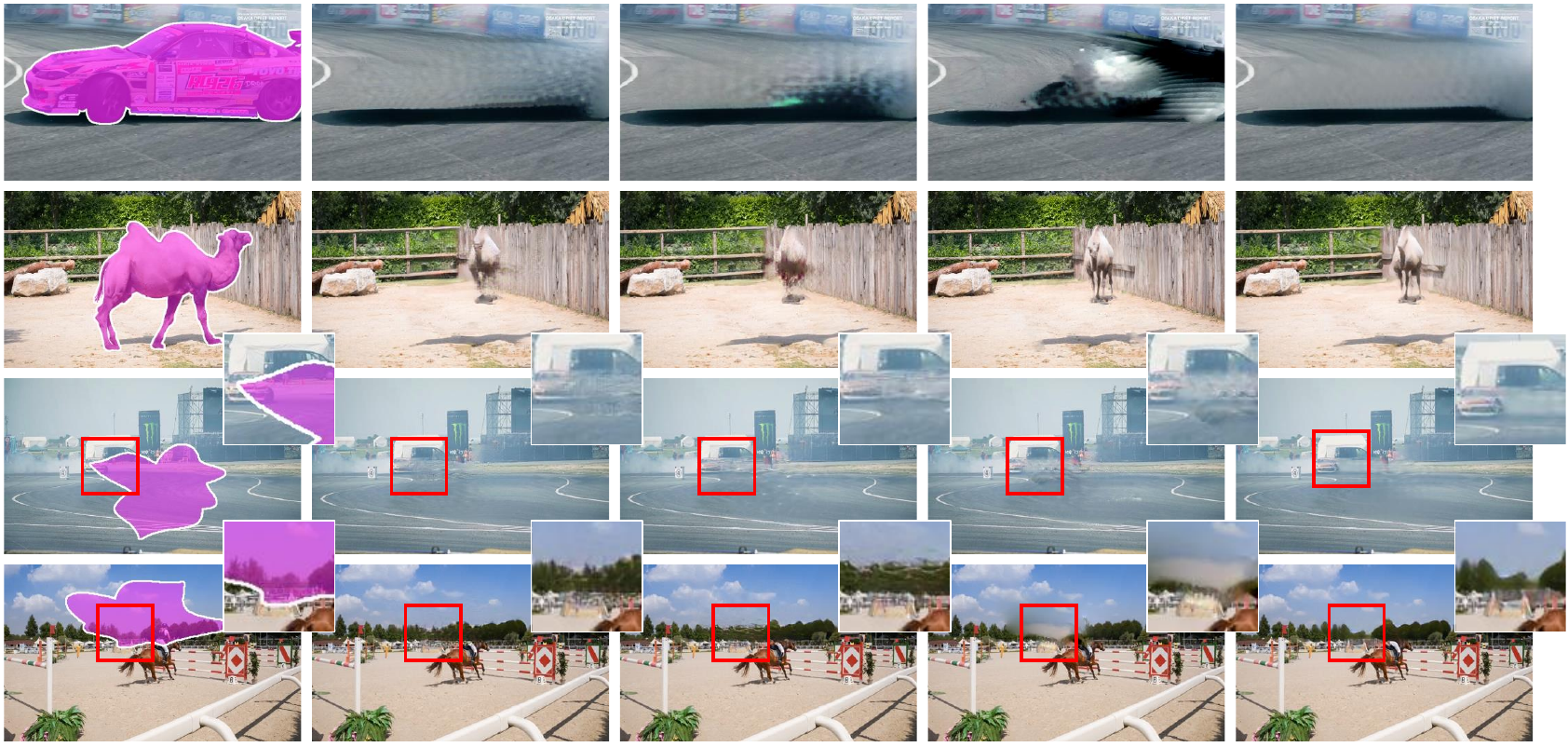}
    \small
    \put(3.7,-1.5){Masked Frames}
    \put(24,-1.5){FuseFormer~\cite{liu21fuseformer}}
    \put(45,-1.5){E$^2$FGVI~\cite{li22e2fgvi}}
    \put(66,-1.5){FGT~\cite{zhang22flow}}
    \put(86.8,-1.5){Ours}
    \end{overpic}
    \vspace{-1mm}
  \caption{Qualitative video inpainting results compared with FuseFormer~\cite{liu21fuseformer}, E$^2$FGVI~\cite{li22e2fgvi}, and FGT~\cite{zhang22flow}.}
  \label{fig:qualitative_video}
  \vspace{-1mm}
\end{figure*}

{\noindent\bf Analysis.} The fundamental issue in continual learning is balancing plasticity and stability~\cite{mermillod2013stability}. Maintaining a copy of the pre-trained model weights can compromise the plasticity of the video model, as verified in the experiments (see Table~\ref{tab:temporal_ii_sota}). In such cases, the model may struggle to adapt to the video domain and become trapped in a local optimum. Therefore, we train DMT${_\text{vid}}$ from scratch instead of fine-tuning it on DMT${_\text{img}}$. The use of ReLU activation in Equation~\ref{eqn:mig} helps suppress negative information while retaining positive information. Similarly, the Hadamard product of the dynamic mask $m$ used in Equation~\ref{eqn:mig} blocks invalid signals in the masked regions. It is worth noting that the mask $m$ is iteratively updated using the Mask Activation algorithm.

\subsection{One-shot Object Removal}

Existing video inpainting research~\cite{li22e2fgvi,zhang22flow,cai22devit} typically relies on frame-wise masks to specify inpainting regions. However, creating frame-wise masks can be labor-intensive, especially for long-term videos in real-world scenarios. Several prior studies have explored inpainting videos without frame-wise masks, opting instead for inpainting guided by a single-frame mask~\cite{ouyang21inter, lee2023one} or a click point~\cite{yang2023track}. To improve the ease of application interaction, we introduce a one-shot object removal pipeline that allows users to provide simple text or stroke inputs.

Building upon our proposed DMT, we incorporate SEEM~\cite{zou2023seem} to handle one-shot input and predict the segmentation of the selected object. As depicted in Figure~\ref{fig:teaser} (c), our method removes unwanted objects from the video by providing a text prompt that corresponds to the unwanted object. Compared to using a single-frame mask or a click point, providing input through text is a simpler way to select the unwanted object. Additionally, we offer stroke-based input, where users can draw a point or brush over the unwanted object in the reference image. For further details, please refer to the video demonstration or the provided demo code.

\section{Experiments}

\noindent
\textbf{Implementation details}. For training the DMT in both the image and video domains, we utilize eight A100 GPUs. For experiments involving quantitative and efficiency comparisons, we use one RTX 3090 GPU. The DMT${_\text{vid}}$ is trained using the Adam optimizer with a batch size of 8 and an initial learning rate of 0.0001. The learning rate is halved at iterations 30e4, 40e4, and 45e4. The DMT${_\text{img}}$ is trained using the Adam optimizer with a batch size of 32 and a learning rate of 0.001. It is important to note that video inpainting requires maintaining temporal consistency, handling referencing between frames, and addressing viewpoint changes, while image inpainting does not. Due to this significant gap between the two tasks, it is common to approach video and image tasks separately. Although some methods~\cite{xu19dfc} attempt to embed image inpainting into video inpainting to handle deficiency cases, the efficiency of such multi-stage approaches remains a concern.

\noindent
\textbf{Dataset}. To evaluate the DMT$_{\text{vid}}$ for video inpainting, we assess its performance on two datasets: YouTube-VOS~\cite{youtubevos} and DAVIS~\cite{davis}. YouTube-VOS contains 3,471 videos for training and 508 videos for testing. Following the approach in~\cite{liu21fuseformer}, we use one hundred video clips from DAVIS for training the model and report the experimental metrics on the remaining 50 video clips.

\noindent
\textbf{Evaluation metric}.
To comprehensively evaluate the performance of our model, we utilize three different metrics: PSNR, SSIM~\cite{DBLP:journals/tip/WangBSS04}, and VFID~\cite{DBLP:conf/nips/Wang0ZYTKC18}. These metrics are commonly used in previous video inpainting literature~\cite{zeng20sttn,liu21fuseformer}. To ensure fairness in evaluation, we employ the same pair of testing images and masks and use the same video sampling process. In video experiments, we use DMT${\text{img}}$ and DMT${\text{vid}}$ both trained on YouTube-VOS to avoid information leakage.

\begin{figure}[t]
  \centering
  \includegraphics[width=\columnwidth]{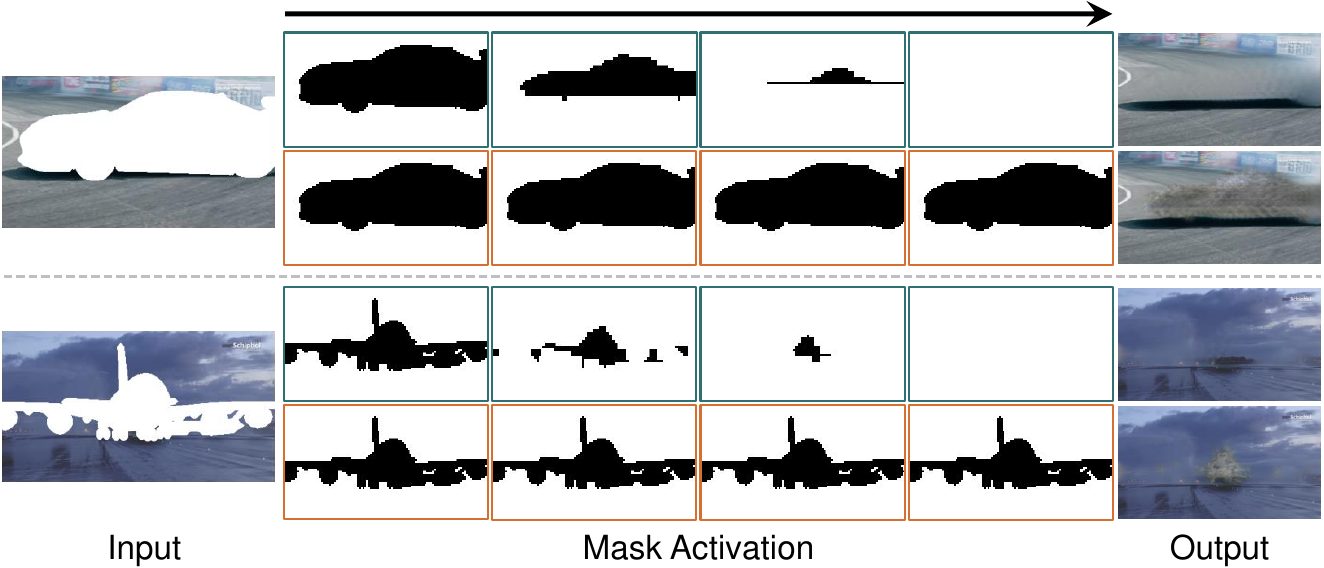}
  \caption{Mask activation effect on qualitative results. The green-line box indicates our mask activation, while the red-line box denotes inference without mask activation.}\label{fig:illustrate_mask_updater}
\end{figure}

\begin{table}[b]
\centering
\small
\caption{Comparison with the SOTA of image inpainting on DAVIS dataset. LAMA~\cite{suvorov22lama} indicates a pre-trained image inpainting model. To activate temporal-awareness, LAMA$\P$ denotes utilizing pre-trained weights and fine tuning on the video dataset with flow-guided features~\cite{li22e2fgvi} and 3D discriminator loss~\cite{zeng20sttn}.}
\label{tab:temporal_ii_sota}
\setlength{\tabcolsep}{7.65pt}
\begin{tabular}{lc|ccc}
\toprule \rowcolor{TableHead}
Method              & \makecell[c]{Temporal \\ Awareness} & PSNR$\uparrow$ & SSIM$\uparrow$ & VFID$\downarrow$ \\ \midrule
LAMA~\cite{suvorov22lama}                & \XSolidBrush        & 30.25 & 0.9560   &  0.181       \\
LAMA$\P$            & \Checkmark          & 32.23	 & 	0.9672 & 0.137         \\ \midrule \rowcolor{RowColor}
DMT$_\text{img}$ & \XSolidBrush        & 28.97 & 0.9475 & 0.162 \\ \rowcolor{RowColor}
DMT$_\text{vid}$ & \Checkmark & 33.82 & 0.9759 & 0.104        \\ \bottomrule        
\end{tabular}
\end{table}

\noindent
\textbf{Baselines}.
We present quantitative findings under a free-form masks setting~\cite{li22e2fgvi} on YouTube-VOS~\cite{youtubevos} and DAVIS~\cite{davis} in Table~\ref{tab:main_video}. We compare our method to existing video inpainting methods such as VINet~\cite{kim19deep}, DFVI~\cite{xu19dfc}, LGTSM~\cite{chang19lgstm}, CAP~\cite{lee19copy}, STTN~\cite{zeng20sttn}, FGVC~\cite{gao20fgvc}, FuseFormer~\cite{liu21fuseformer}, E$^2$FGVI~\cite{li22e2fgvi}, and FGT~\cite{zhang22flow}. E$^2$FGVI and FGT are the SOTAs in video inpainting based on optical flow and transformer. For additional settings of masks, please refer to the supplementary video inpainting results.

\subsection{Results Analysis}

\noindent
\textbf{Comparison with state-of-the-art methods}. Our approach outperforms all previous state-of-the-art models on all three quantitative metrics, as illustrated in Table~\ref{tab:main_video}. The superior results demonstrate how our method produces content that is more faithful to the original frames (PSNR, SSIM) with less distortion (VFID), showcasing the effectiveness of our approach. Figure~\ref{fig:qualitative_video} provides qualitative comparisons against other transformer-based methods~\cite{liu21fuseformer,li22e2fgvi} and flow-guided methods~\cite{li22e2fgvi,zhang22flow}. Notably, FuseFormer~\cite{liu21fuseformer} and E$^2$FGVI~\cite{li22e2fgvi} fail to synthesize the entire body of the distant camel in the first row, while our method produces more refined outcomes with less repetitive mosaic in the second and third rows. The effectiveness of our method in propagating objects, such as the mallard and stroller in the second and fourth rows of Figure~\ref{fig:more_vi_compare}, can also be observed. For additional cases, please refer to the video demonstration.

\noindent
\textbf{From image to video}. To verify the hypothesis that existing image inpainting methods struggle with the video domain, we utilize the state-of-the-art image inpainting method LAMA~\cite{suvorov22lama}, which has shown strong performance in object removal, to inpaint masked videos. For a fair comparison, we introduce a variant called LAMA$\P$, which fine-tunes a pre-trained LAMA model on the video dataset. To enable temporal awareness of LAMA$\P$, we use flow-guided features~\cite{li22e2fgvi} and a 3D discriminator loss~\cite{zeng20sttn}. As shown in Table~\ref{tab:temporal_ii_sota}, LAMA~\cite{suvorov22lama} initially outperforms our DMT$_\text{img}$ in terms of PSNR and SSIM in video inpainting. However, LAMA$\P$, which finetunes the pre-trained image model LAMA~\cite{suvorov22lama}, converges to a local optimum due to weak model plasticity~\cite{mermillod2013stability}. In contrast, our DMT$_\text{vid}$ is trained with Migration Regularization and DMT$_\text{img}$ from scratch, striking a good balance between plasticity and stability and significantly outperforming LAMA$\P$ in all metrics. This verifies the effectiveness of our method in learning an inpainting model from image to video. Additionally, the comparison between the two image models, LAMA~\cite{suvorov22lama} and DMT$_\text{img}$, reveals that LAMA performs better in terms of similarity-wise metrics (PSNR and SSIM), while our DMT$_\text{img}$ performs better in terms of VFID due to its training as a generative model~\cite{li22mat,zhao21comod} to handle large corruptions.

\begin{figure}[t]
  \centering
\includegraphics[width=\columnwidth]{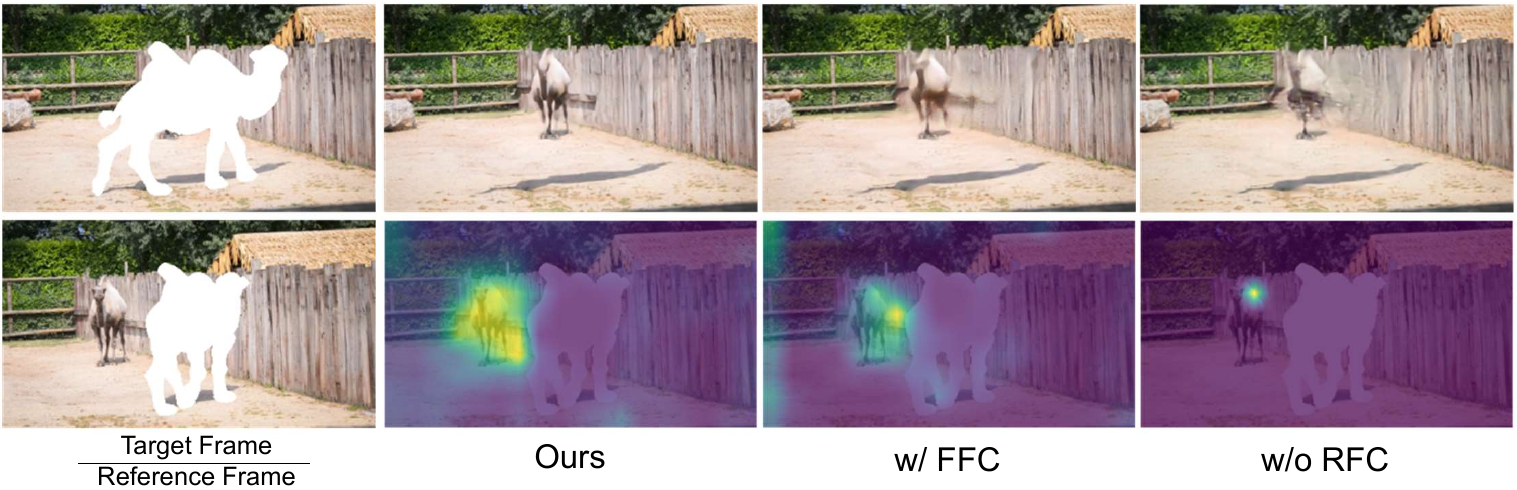}
  \caption{Attention map visualization for ablation experiments on RFC. ``w/ FFC" represents replacing our RFC with fast fourier convolution in LAMA~\cite{suvorov22lama}.}\label{fig:rfc_study}
\end{figure}

\begin{table}[b]
\centering
\caption{Efficiency analysis of unmasked transformer vs. masked transformer. The variant \fullamount{} denotes the transformer layer without dropping invalid tokens. Two window strategies~\cite{yang21focal} are employed to accelerate the vanilla transformer. The variant Valid-Only $x\%$ represents our masked transformer layer with an average input mask ratio of $x$ percent.}
\label{tab:speed_attention}
\small
\setlength{\tabcolsep}{2.9pt}
\begin{tabular}{l|c|cc}
\toprule \rowcolor{TableHead}
& MACs & \multicolumn{2}{c}{Latency$\downarrow$} \\ \rowcolor{TableHead}
Method                     & (G) & \# input frame = 1 & \# input frame = 8  \\ \midrule 
\fullamount{}                       & 6.04 & 0.58 ± 0.071  &  7.83 ± 0.09     \\
\fullamount{} w/ Focal~\cite{yang21focal}                & 6.14 & 1.97 ± 0.133 &  5.42 ± 0.199      \\ \midrule
\rowcolor{RowColor}
Valid-Only 10\%                    & 5.72 & 0.56 ± 0.068  & 5.91 ± 0.236 \\ \rowcolor{RowColor}
Valid-Only 30\%                    & 4.50  & 0.58 ± 0.161 & 3.68 ± 0.193    \\ \rowcolor{RowColor}
Valid-Only 60\%                    & 2.99 & 0.55 ± 0.048 & 1.82 ± 0.104   \\ \rowcolor{RowColor}
Valid-Only 90\%                    & 1.19 & 0.54 ± 0.052 & 0.63 ± 0.093    \\
 \bottomrule
\end{tabular}
\vspace{-2mm}
\end{table}

\noindent
\textbf{Efficiency of Deficiency-aware Masked Transformer}. Our masked transformer effectively reduces computational complexity by dropping invalid tokens before performing multi-head self-attention. In Table~\ref{tab:speed_attention}, we compare the computational complexity (MACs), and running speed (Latency) of self-attention with different schemes. We adopt a focal window attention strategy~\cite{yang21focal} to accelerate self-attention. As shown in Table~\ref{tab:speed_attention}, our masked transformer layer exhibits decreased computational cost and increased inference speed as the input mask ratio increases. Our proposed module still achieves comparable efficiency compared to the vanilla transformer. 


\noindent
\textbf{Evaluation of Receptive Field Contextualizer}. To evaluate the effectiveness of our RFC, we conduct ablation experiments as shown in Table~\ref{tab:abla_largeks} on the DAVIS dataset and train the model for 50,000 iterations (10\% of the complete configuration) on YouTube-VOS~\cite{youtubevos}. We construct variants with different kernel sizes $K$ and a variant that uses Fast Fourier Convolution (FFC)~\cite{suvorov22lama} as a substitute. FFC is a similar module with a high receptive field. Table~\ref{tab:abla_largeks} quantitatively demonstrates that a larger receptive field enhances performance, and our RFC substantially improves PSNR, SSIM, and VFID scores. Figure~\ref{fig:rfc_study} provides a quantitative comparison of our RFC with FFC and no RFC, showcasing the superior performance of our RFC in terms of restoring high-frequency details and enhancing spatiotemporal modeling. 

\begin{table}[t]
\centering
\small
\caption{Investigation on the kernel size $K$ of RFC.}
\label{tab:abla_largeks}
\begin{tabular}{ll|lll}\toprule \rowcolor{TableHead}
Method & \# $K$ & PSNR$\uparrow$ & SSIM$\uparrow$ / \% & VFID$\downarrow$  \\ \midrule \rowcolor{RowColor}
RFC    & 31  & 29.88 & 94.90 & 0.207 \\\rowcolor{RowColor}
RFC    & \textbf{13}  & \textbf{30.26}\textcolor{yyellow}{$_{ \uparrow.16}$} & \textbf{95.35}\textcolor{yyellow}{$_{ \uparrow.27}$} & \textbf{0.188}\textcolor{yyellow}{$_{ \downarrow.007}$} \\ \rowcolor{RowColor}
RFC    & 7   & 30.06 & 95.03 & 0.197 \\ \rowcolor{RowColor}
RFC    & 3   & 29.79 & 94.76 & 0.210 \\ \rowcolor{RowColor}
RFC    & 1   & 29.72 & 94.65 & 0.213 \\ \midrule
w/ FFC~\cite{suvorov22lama}    & -   & \underline{30.10} & \underline{95.08} & \underline{0.195} \\
w/o RFC & - & 29.78 & 94.77 & 0.219 \\  \bottomrule
\end{tabular}
\end{table}

\subsection{Ablation Study}

In this section, we conduct an ablation study to gauge the contributions of the proposed components in our framework and verify our hypothesis. Table~\ref{tab:ablation} provides an overview of the contribution of each component of DMT based on experiments conducted on the DAVIS dataset. Based on the analysis of the PSNR, SSIM, and VFID metrics, our RFC and Migration Regularization provide key improvements in video inpainting performance.

\noindent
\textbf{Effectiveness of Token Selection}. Our proposed Token Selection mechanism, which explicitly discards masked tokens, helps reduce invalid noise and computational complexity. The results in Table~\ref{tab:speed_attention} demonstrate the higher inference efficiency achieved by this mechanism.

\noindent
\textbf{Effectiveness of Mask Activation}. We find that DMT without Mask Activation leads to significant performance degradation, as it becomes challenging to reconstruct invalid tokens solely relying on the convolutional Decoder. Figure~\ref{fig:illustrate_mask_updater} visually illustrates the process of mask activation, highlighting the blurred results in the second and fourth rows that indicate the ineffectiveness of our method in performing effective spatiotemporal correlation modeling without mask activation.

\noindent
\textbf{Effectiveness of RFC}. Omitting the Receptive Field Contextualizer from the network training results in a loss of high-frequency details and positional information, as discussed in Section~\ref{sec:dmt}. The results in Table~\ref{tab:abla_largeks} demonstrate that incorporating our RFC module significantly enhances all three novelty scores, indicating its ability to capture high-frequency signals and learn positions. We empirically choose a large kernel size of $K=13$ for the RFC.

\noindent
\textbf{Effectiveness of Migration Regularization}. By applying the Migration Regularization in Equation~\ref{eqn:mig} to the network, we can enhance video inpainting performance by transferring the generative prior of a pre-trained image inpainting model without requiring direct access to the image dataset. Moreover, this shows that our Migration Regularization facilitates new tasks with old domain knowledge, achieving a good balance between plasticity and stability. Please refer to the supplementary material for inpainting comparisons with large corruptions.

\begin{table}[t]
\centering
\caption{Exploration of different component in DMT. }
\label{tab:ablation}
\small
\begin{tabular}{llll}
\toprule \rowcolor{TableHead}
Method   & PSNR$\uparrow$ & SSIM$\uparrow$ / \% & VFID$\downarrow$  \\ \midrule \rowcolor{RowColor}
Ours    & 33.82       & 97.59 & 0.104    \\ \midrule
w/o token selection   & 33.74\textcolor{othercolor}{$_{ \downarrow0.08}$}
& 97.53\textcolor{othercolor}{$_{\downarrow0.05}$}
& 0.106\textcolor{othercolor}{$_{\uparrow0.002}$}    \\
w/o mask activation & 31.98\textcolor{othercolor}{$_{ \downarrow1.84}$} &
	96.41\textcolor{othercolor}{$_{ \downarrow1.18}$} &
	0.142\textcolor{othercolor}{$_{ \uparrow0.038}$}
   \\
w/o RFC & 33.43\textcolor{othercolor}{$_{ \downarrow0.39}$} &
	97.39\textcolor{othercolor}{$_{ \downarrow0.2}$} &
	0.109\textcolor{othercolor}{$_{ \uparrow0.005}$}
   \\
w/o $\mathcal{L}_\text{mig}$ & 33.51\textcolor{othercolor}{$_{ \downarrow0.31}$} &
	97.47\textcolor{othercolor}{$_{ \downarrow0.12}$} &
	0.108\textcolor{othercolor}{$_{ \uparrow0.004}$}
   \\

 \bottomrule
\end{tabular}
\end{table}

\section{Conclusion}

In conclusion, we have introduced a novel Deficiency-aware Masked Transformer for video inpainting. Our proposed Migration Regularization effectively enables the model to handle deficiency cases. The components of Token Selection, Mask Activation, and Receptive Field Contextualizer have all made significant contributions to the overall performance improvement.
Extensive experiments have demonstrated the superiority of our method over SOTAs in terms of quantitative metrics and visual quality. Our DMT exhibits robust generalization to in-the-wild input and easily adapts to one-shot object removal task, showcasing its potential for various applications in video editing and restoration.
To the best of our knowledge, we are the first to propose leveraging pre-trained image inpainting models for video inpainting. We believe that our approach will inspire further research in connecting image inpainting with video inpainting and vice versa.


{\small
\bibliographystyle{ieee_fullname}
\bibliography{egbib}
}

\clearpage
\section{Appendix}
\begin{figure*}[t]
  \centering
    \begin{overpic}[width=\textwidth]{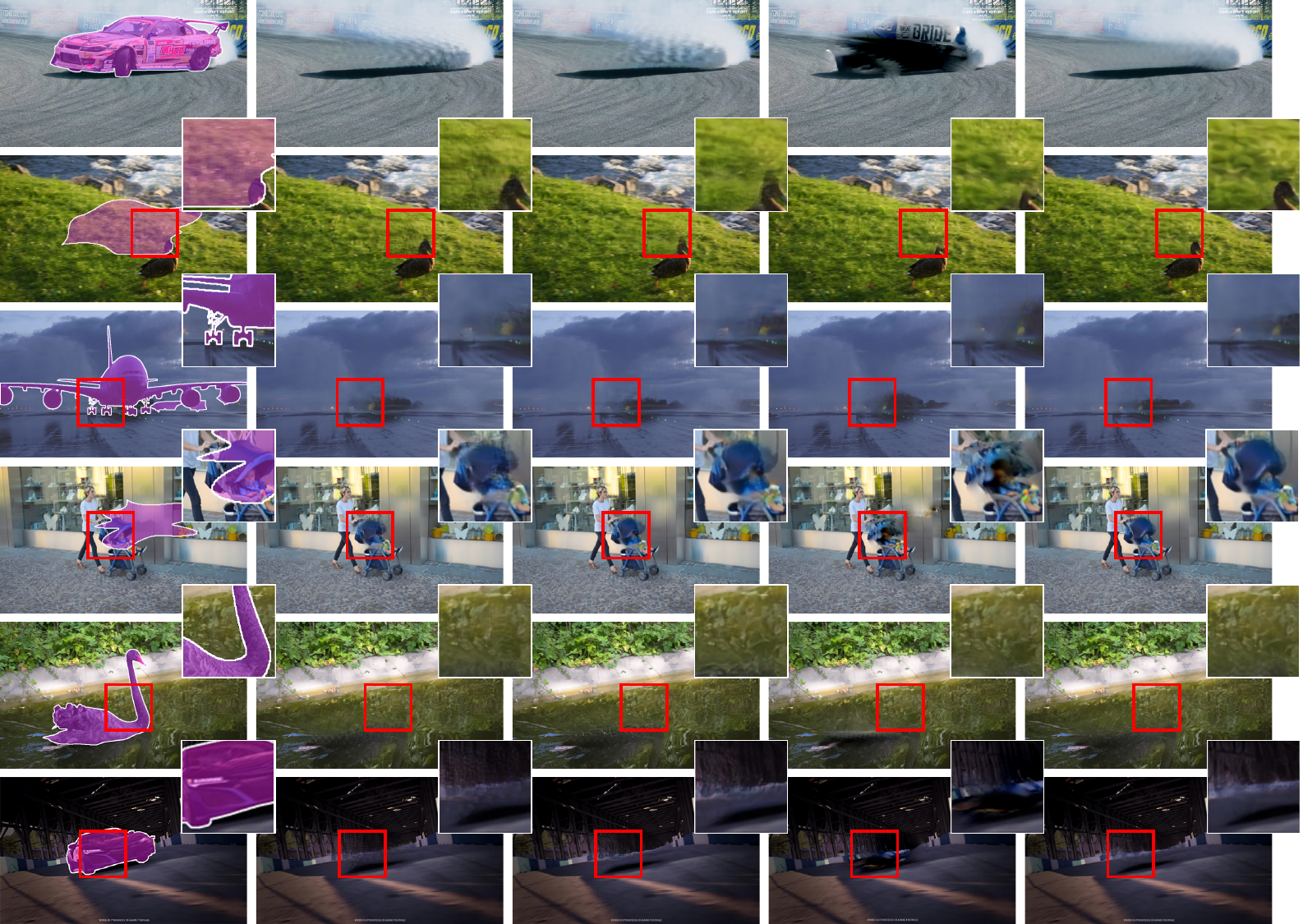}
    \small
    \put(3.7,-1.5){Masked Frames}
    \put(24,-1.5){FuseFormer~\cite{liu21fuseformer}}
    \put(45,-1.5){E$^2$FGVI~\cite{li22e2fgvi}}
    \put(66,-1.5){FGT~\cite{zhang22flow}}
    \put(86.8,-1.5){Ours}
    \end{overpic}
    \vspace{-1mm}
  \caption{
More qualitative results compared to the state-of-the-art video inpainting methods FuseFormer~\cite{liu21fuseformer}, E$^2$FGVI~\cite{li22e2fgvi}, and FGT~\cite{zhang22flow}. \emph{Best viewed in color and by zooming in for all figures throughout the supplementary.}
}
  \label{fig:more_vi_compare}
  \vspace{-1mm}
\end{figure*}


\subsection{Token Selection Complexity Analysis}
Due to reducing the number of tokens before patch embedding and self-attention, our masked transformer lessens the computational strain quadratically. Formally, the varying time complexity from the standard transformer to the masked transformer is represented as $O(4Nd^2 + 2N^2d) \rightarrow O(4N^\prime d^2 + 2{N^\prime}^2d)$, where $d$ indicates the feature dimension of embedded tokens. And the number of tokens of the masked transformer $N^\prime$ subjects to: 
\begin{equation}
    N^\prime < N, \quad N^\prime \sim N \cdot \frac{\mathbbm{1}(m)}{|m|},
\end{equation}
where $\mathbbm{1}(\cdot)$ stands for the number of elements whose value is equal to $1$, \ie, masked transformers keep complexity inversely proportional to mask coverage.

\subsection{Evaluation on More Mask Settings}

The performance of video inpainting varies for random masks and random videos. 
In addition to the free-form mask setting~\cite{li22e2fgvi} used in the paper, we follow an existing method~\cite{zou2021progressive} to evaluate the video inpainting task in a formulaic setting. To assess the performance of our approach, we conduct quantitative experiments on the DAVIS dataset using three types of masks - Curve, Stationary, and Object. We present our results and compare them with the state-of-the-art E$^2$FGVI~\cite{li22e2fgvi}, FGT~\cite{zhang22flow}, and FuseFormer~\cite{liu21fuseformer} methods in Tab.~\ref{tab:compare_formulaic}.

\begin{figure*}[t]
  \centering
    \begin{overpic}[width=\textwidth]{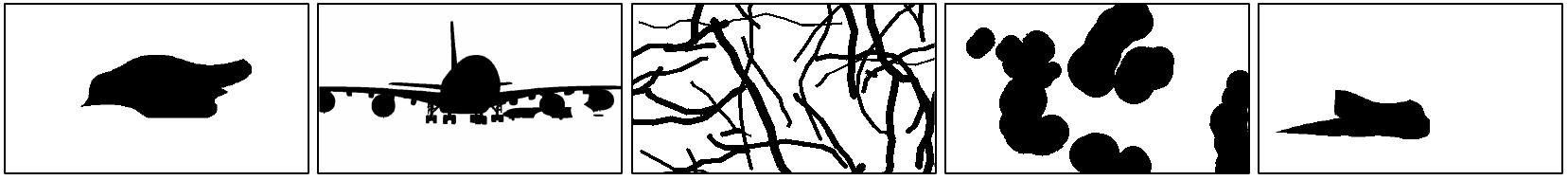}
    \small
    \put(7,-1.5){Free form}
    \put(25,-1.5){Segmentation}
    \put(47,-1.5){Curve~\cite{zou2021progressive}}
    \put(67,-1.5){Object~\cite{zou2021progressive}}
    \put(85,-1.5){Stationary~\cite{zou2021progressive}}
    \end{overpic}
    \vspace{-1mm}
  \caption{The examples of mask settings in our experimental evaluation.}
  \label{fig:vid_masks}
  \vspace{-1mm}
\end{figure*}

\begin{figure*}[t]
  \centering
  \includegraphics[width=\textwidth]{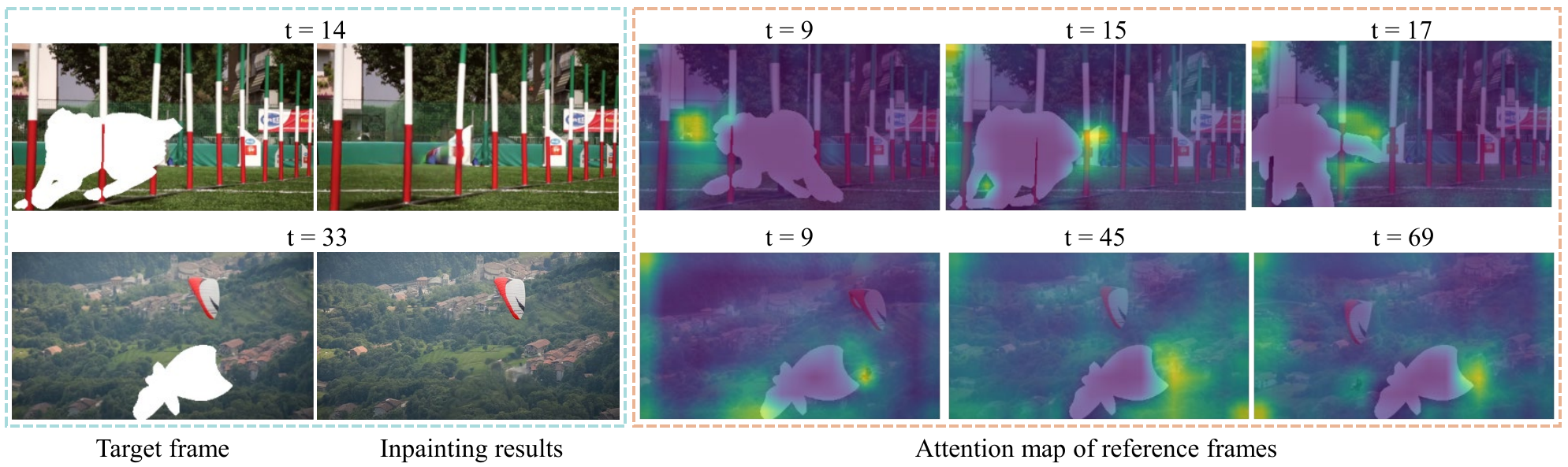}
  \vspace{-5mm}
  \caption{Illustration of the cross-frame attention maps learned by DMT for missing regions of the target frame. The color gradient, ranging from blue to green to yellow, represents the gradual increase in attention values.}\label{fig:attention_map}
  \vspace{-3mm}
\end{figure*}

\renewcommand\arraystretch{0.8}
\begin{table}[h]\scriptsize
  \centering
  \caption{Comparison in the formulaic mask settings.}
\begin{tabular}{l|ccc}
           & \multicolumn{3}{c}{PSNR$\uparrow$/SSIM$\uparrow$/FID$\downarrow$}                            \\ \cline{2-4} 
           & \textbf{Curve Mask}              & \textbf{Stationary Mask}         & \textbf{Object Mask}             \\ \hline
Ours       & \textbf{35.77}/\textbf{0.991}/\textbf{0.122} & \textbf{37.14}/\textbf{0.982}/\textbf{0.084} & \textbf{27.39}/\textbf{0.942}/\textbf{0.252} \\
E$^2$FGVI  & 34.58/0.988/0.154 & 36.52/0.979/0.089 & 26.97/0.934/0.259 \\
FuseFormer &    33.34/0.984/0.201                & 35.85/0.977/0.102                & 26.04/0.920/0.308                  
\end{tabular}
\label{tab:compare_formulaic}
\end{table}%

As demonstrated in Table~\ref{tab:compare_formulaic}, our method outperforms state-of-the-art approaches in all formulaic settings w.r.t. PSNR, SSIM, and VFID metrics. These results showcase the superior performance of our method across various experimental settings and demonstrate its better generalization capability.

To illustrate the different masks used in our experiments, we provide examples in Fig.~\ref{fig:vid_masks}. We present the masks from right to left, starting with free-form, segmentation, curve, object, and stationary masks. The segmentation mask is a manually crafted segmentation label from the dataset. During the training of our DMT$_{vid}$ model, we used randomly generated free-form masks. We evaluated the performance of our approach using both free-form masks and segmentation masks through qualitative comparisons.

\subsection{Long-range Dependencies Modeling}

Our proposed framework leverages the power of DMT to model long-range dependencies and exploit frame-to-frame correlations. Fig.~\ref{fig:attention_map} shows the attention maps learned by DMT$_\text{vid}$ in the last layer. For instance, when a running dog in a video is occluded by a segmentation mask (t=9, 15, 17), our model fills the missing regions with coherent background texture. Similarly, in a video with a man on a parachute partially corrupted by a random mask, our model accurately tracks the flying person across frames.

\subsection{More Qualitative Results}
 
Figure~\ref{fig:more_vi_compare} shows additional qualitative results of our video inpainting method and compares it with three state-of-the-art methods: E$^2$FGVI~\cite{li22e2fgvi}, FGT~\cite{zhang22flow}, and FuseFormer~\cite{liu21fuseformer}. Our method outperforms them in several aspects: it produces less ghosting artifacts (see row four), less mosaic or repeated patterns (see rows one and six), more complete foreground objects (see row two), and can handle object masks of various scales (see rows one, three, five and six).

\begin{figure*}[t]
  \centering
  \includegraphics[width=\textwidth]{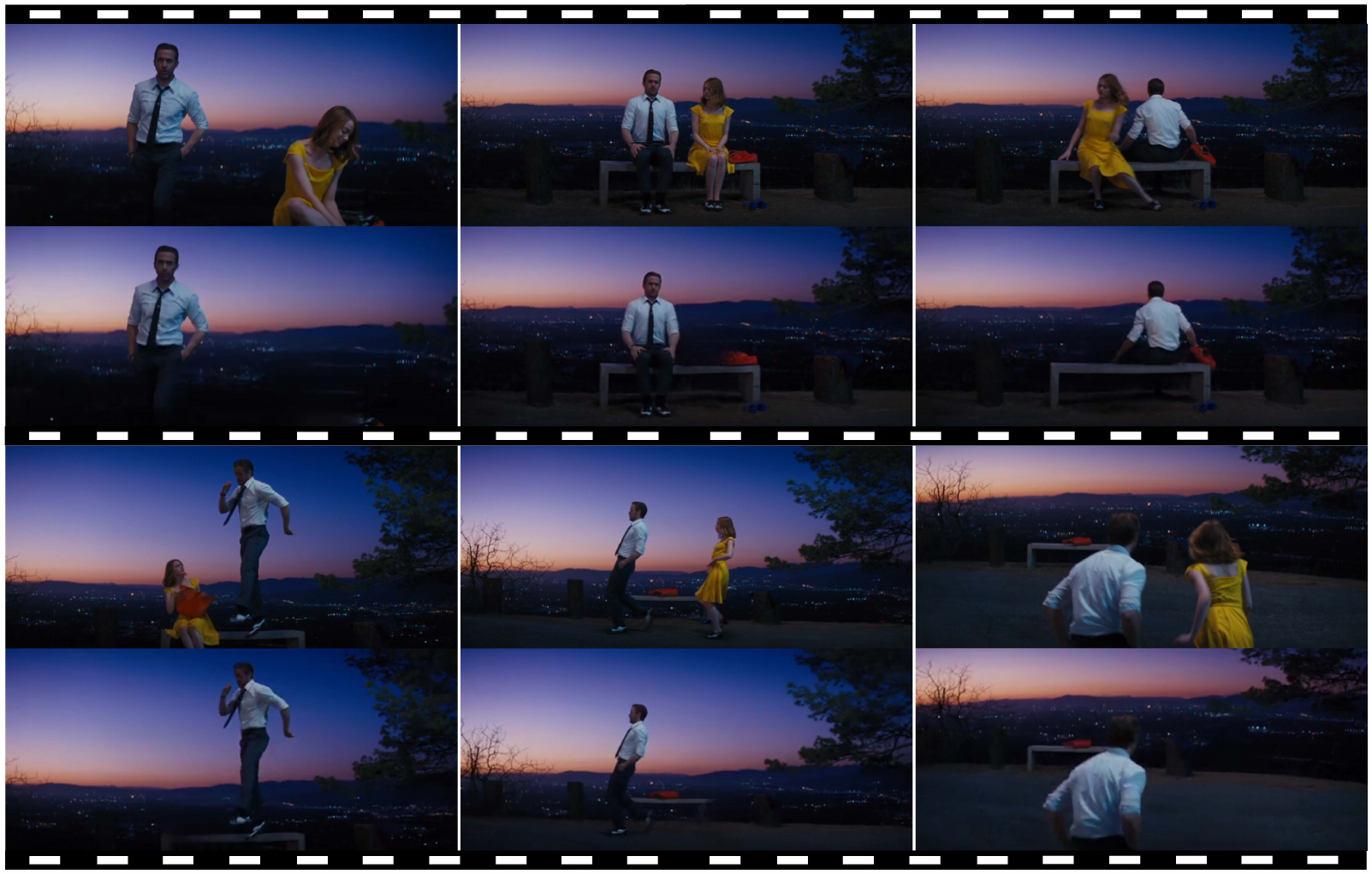}
  \vspace{-5mm}
  \caption{Long-term video inpainting result comprising 1358 frames, inferring with the text prompt "The female dancer wearing a yellow outfit" without using frame-wise masks. For a video demonstration, please refer to our code repository.}\label{fig:longterm_or}
  \vspace{-3mm}
\end{figure*}

\begin{figure*}[t]
  \centering
  \includegraphics[width=0.9\textwidth]{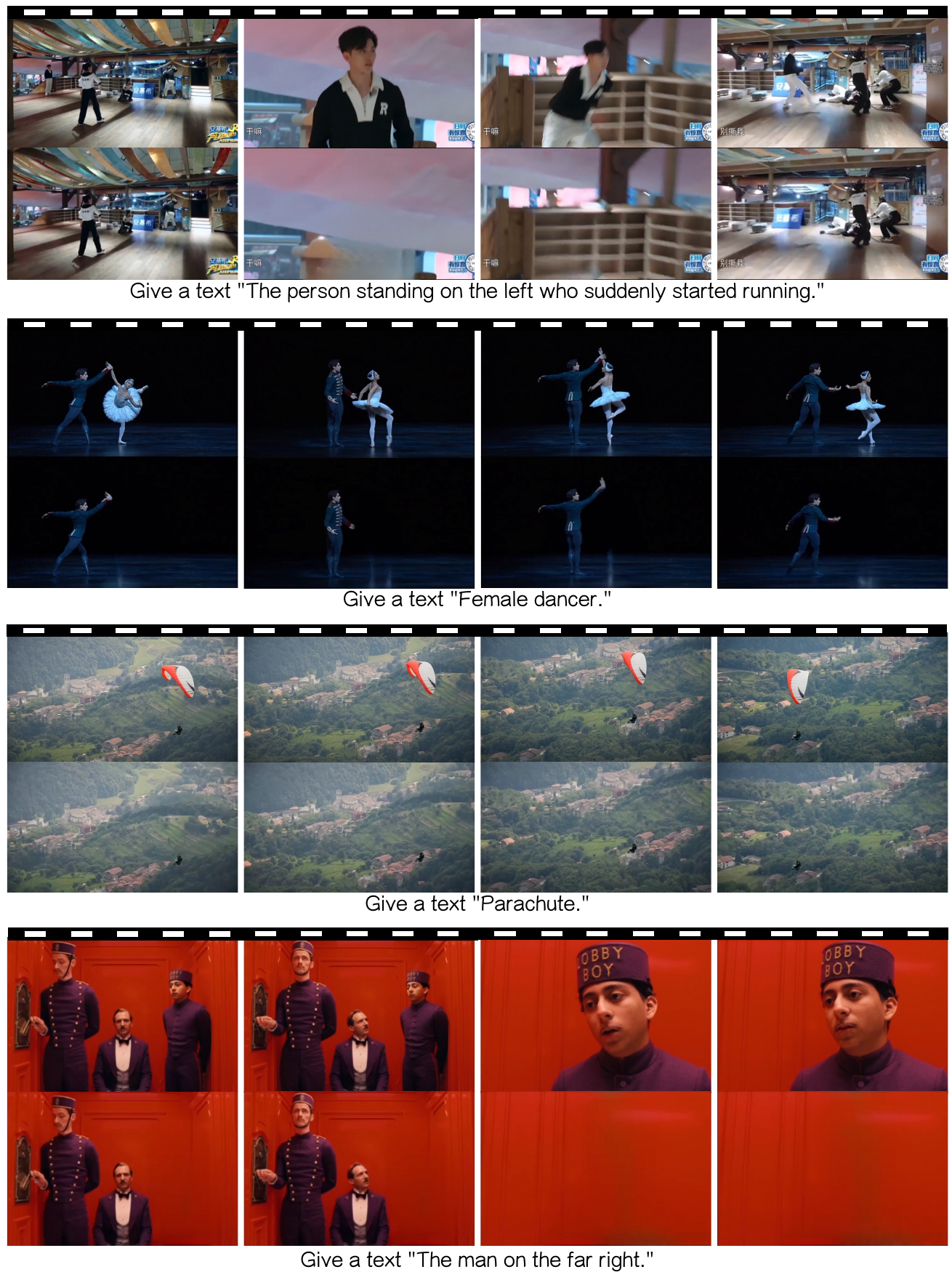}
  \caption{Results of the text-guided object removal task. For a complete video demonstration, please visit our code repository.}\label{fig:textguide_or}
  \vspace{-3mm}
\end{figure*}

Our model demonstrates strong performance in various video inpainting tasks, as shown in Figure~\ref{fig:longterm_or} for long-term video inpainting and Figure~\ref{fig:textguide_or} for text-guided object removal. Notably, our method achieves effective inpainting of unwanted objects without visible artifacts and generalizes to in-the-wild and one-shot video inpainting tasks without the need for frame-wise masks.

\subsection{By-product Image Inpainting}

In addition to the video inpainting model, we obtain an image inpainting model DMT$_{\text{img}}$ as a by-product. We employ Places2~\cite{places2} as train and test dataset for the image inpainting experiments. For evaluating recent image inpainting methods, we employ perceptual metrics including Frechet Inception Distance (FID)~\cite{DBLP:conf/nips/HeuselRUNH17}, Paired/Unpaired Inception Discriminative Score (P/U-IDS)~\cite{zhao21comod}. 

We compare our model with state-of-the-art image inpainting methods, including DeepFill~\cite{yu19free}, Lama~\cite{suvorov22lama}, CoModGAN~\cite{zhao21comod}, MISF~\cite{li22misf}, TFill~\cite{zheng22tfill}, and FcF~\cite{jain22fcf}. The extensive image completion comparisons are carried out on Places2 dataset in terms of random masks with small and large coverage ratios. Mask ratios in the Small and Large settings, respectively, roughly range from 10\% to 30\% and 40\% to 90\%. From Table~\ref{tab:main_places2}, our approach shows the best performance with respect to FID, P-IDS, and U-IDS quantitative metrics on Places2 dataset.

To further study our DMT$_\text{img}$, we provide qualitative comparison results on the DAVIS dataset. 
The reconstruction results of Lama~\cite{suvorov22lama}, CoModGAN~\cite{zhao21comod}, TFill~\cite{zheng22tfill}, and FcF~\cite{jain22fcf} are presented for comparisons.
Compared to these advanced image inpainting approaches, our DMT$_\text{img}$ can produce more accurate textural and structural data and more consistent content in masked areas, as shown in Fig.~\ref{fig:qualitative_image_places}. Moreover, in cases of large corruptions, our DMT$_\text{img}$ can reconstruct the image content with realistic semantics instead of repeating patterns caused by inherent ambiguity.

\begin{table}[t]
\small
\caption{Quantitative evaluation of image inpainting methods on Places dataset~\cite{places2}.}
\label{tab:main_places2}
\centering
\setlength{\tabcolsep}{2.8pt}
\begin{tabular}{lcccccc}
\toprule 
            & \multicolumn{2}{c}{FID$\downarrow$} & \multicolumn{2}{c}{P-IDS$\uparrow$} & \multicolumn{2}{c}{U-IDS$\uparrow$} \\ 
Method      & Small            & \multicolumn{1}{c|}{Large}            & Small            & \multicolumn{1}{c|}{Large}            & Small            & Large            \\ \cmidrule(r){1-1} \cmidrule(r){2-3} \cmidrule(r){4-5} \cmidrule{6-7}
DeepFill~\cite{yu19free}    & 2.598           & 21.403          & 0.055           & 0.006           & 0.304           & 0.098 \\
CoModGAN~\cite{zhao21comod}    & 1.177           & 6.680           & 0.200           & 0.087           & 0.396           & 0.251           \\
MISF~\cite{li22misf}        & 4.458           & 17.451          & 0.003           & 0.001           & 0.126           & 0.064           \\
Lama~\cite{suvorov22lama}        & \underline{0.910}           & 8.331           & 0.193           & 0.045           & 0.402           & 0.219           \\
TFill~\cite{zheng22tfill}       & 1.417           & 14.537          & 0.126           & 0.021           & 0.361           & 0.159           \\
FcF~\cite{jain22fcf}         & \underline{0.910}           & \underline{5.550}           & \underline{0.285}           & \underline{0.132}           & \underline{0.426}           & \underline{0.289}           \\
\textbf{DMT$_{\text{img}}$} & \textbf{0.766}  & \textbf{4.208}  & \textbf{0.300}  & \textbf{0.156}  & \textbf{0.434}  & \textbf{0.314}  \\
\bottomrule         
\end{tabular}
\end{table}

\subsection{Limit Dicussion}

We have introduced a novel method with promising results for video inpainting. However, our approach does have two main limitations.
Firstly, while our method exhibits robustness to random video shapes, one common drawback of Transformers-based methods~\cite{liu21fuseformer,zeng20sttn,li22e2fgvi} is their high memory requirements when processing high-resolution content, such as 1080p videos. The attention matrix product necessitates significant memory resources, particularly for videos with large temporal scales or resolutions.
Secondly, we observed through empirical studies that utilizing a unified model trained on both image and video datasets hinders achieving state-of-the-art performance in both tasks. This discrepancy arises from the substantial gap between image and video inpainting objectives: image inpainting focuses on generatively filling missing regions, while video inpainting requires referencing cross-frame information and maintaining temporal coherence. This highlights the ongoing challenge in addressing both image and video inpainting in such scenarios.

\newcommand{\MMM}{0.155}
\renewcommand\arraystretch{0.8}
\begin{figure*}[t]
\centering
\begin{tabular}{c@{\hspace{1mm}}c@{\hspace{1mm}}c@{\hspace{1mm}}c@{\hspace{1mm}}c@{\hspace{1mm}}c}
\small{Lama} & \small{CoModGAN} & \small{TFill} & \small{FcF} & \small{Ours} & \small{Masked} \\

\frame{\includegraphics[width=\MMM\linewidth]{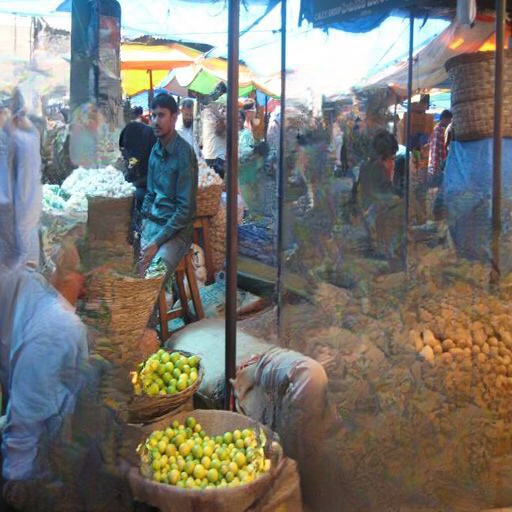}} &
\frame{\includegraphics[width=\MMM\linewidth]{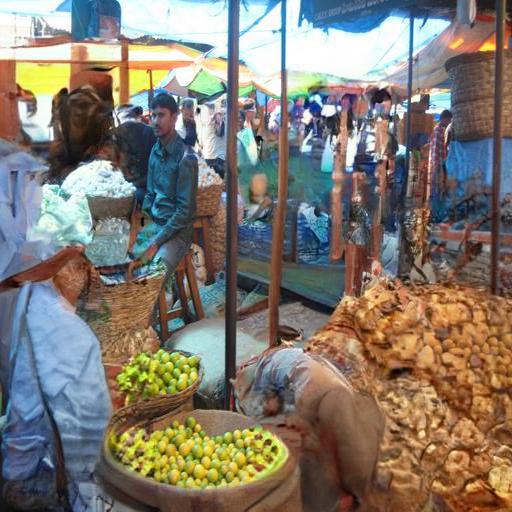}} &
\frame{\includegraphics[width=\MMM\linewidth]{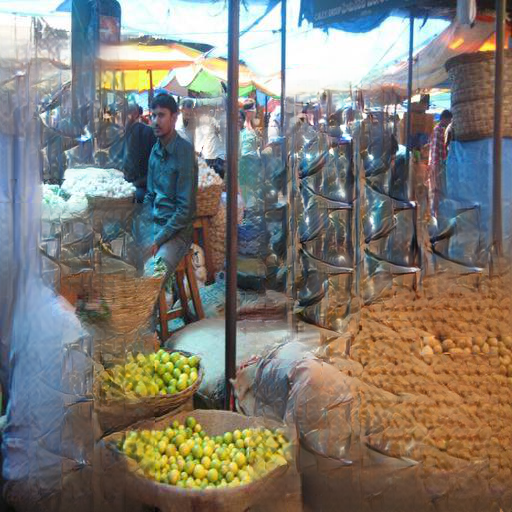}} &
\frame{\includegraphics[width=\MMM\linewidth]{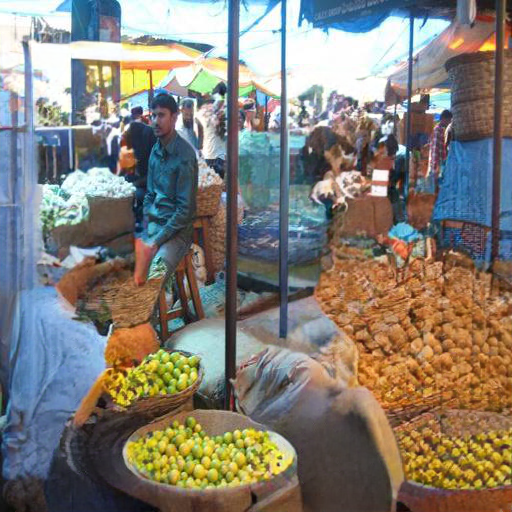}} &
\frame{\includegraphics[width=\MMM\linewidth]{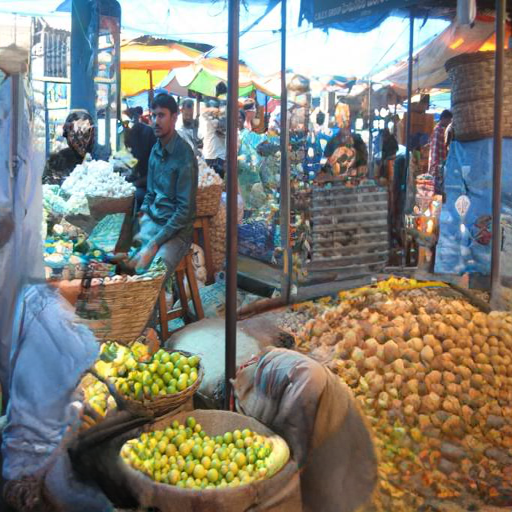}} &
\frame{\includegraphics[width=\MMM\linewidth]{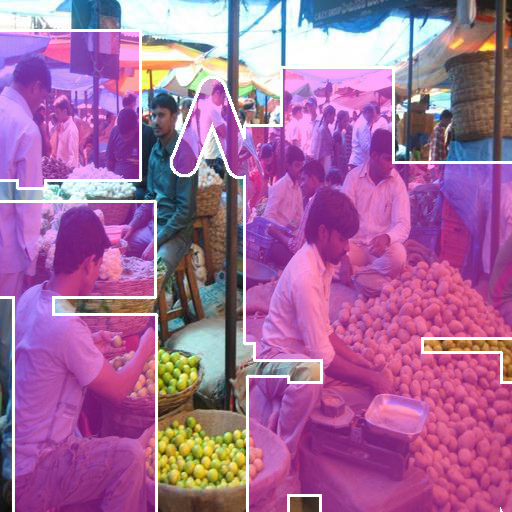}} \\

\frame{\includegraphics[width=\MMM\linewidth]{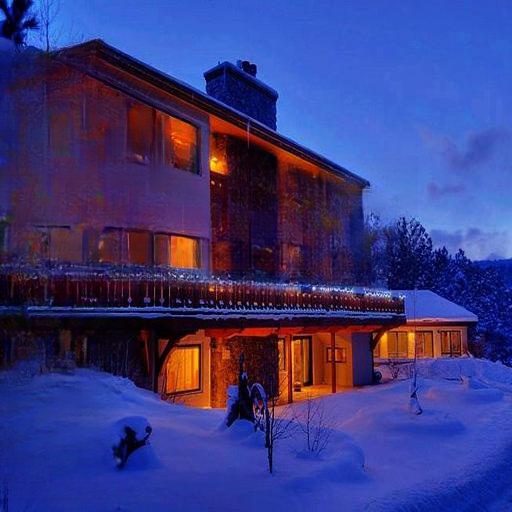}} &
\frame{\includegraphics[width=\MMM\linewidth]{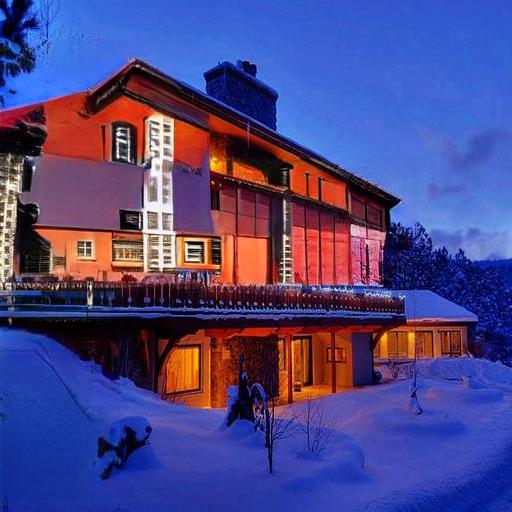}} &
\frame{\includegraphics[width=\MMM\linewidth]{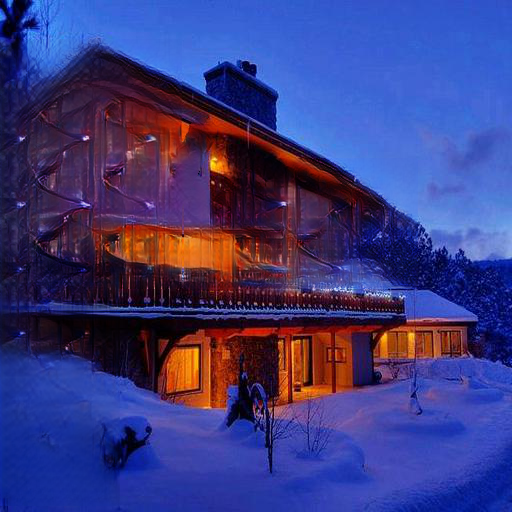}} &
\frame{\includegraphics[width=\MMM\linewidth]{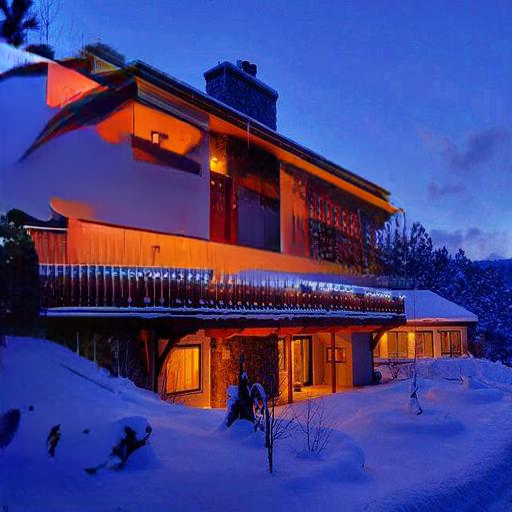}} &
\frame{\includegraphics[width=\MMM\linewidth]{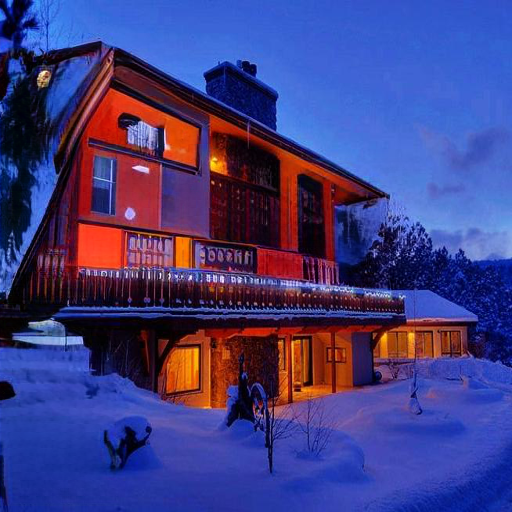}} &
\frame{\includegraphics[width=\MMM\linewidth]{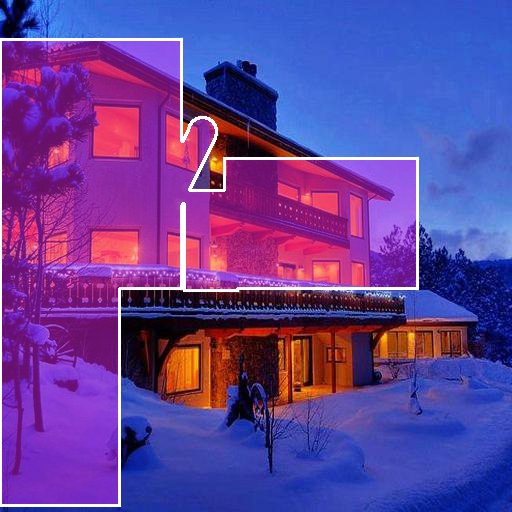}} \\

\frame{\includegraphics[width=\MMM\linewidth]{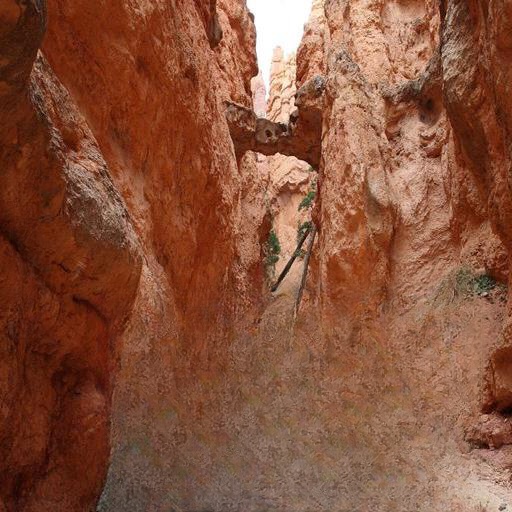}} &
\frame{\includegraphics[width=\MMM\linewidth]{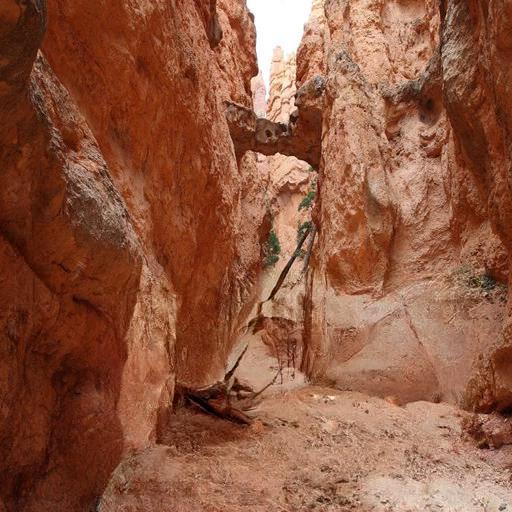}} &
\frame{\includegraphics[width=\MMM\linewidth]{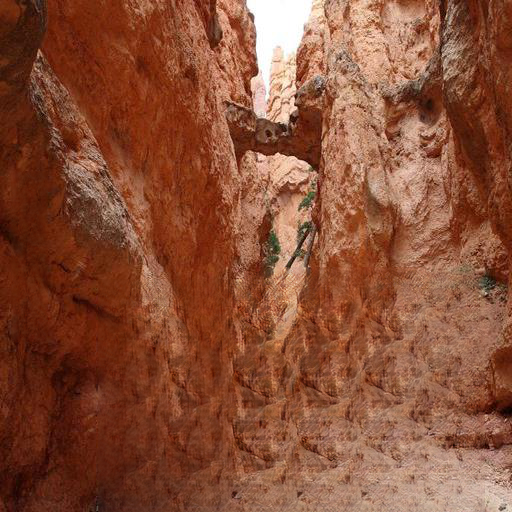}} &
\frame{\includegraphics[width=\MMM\linewidth]{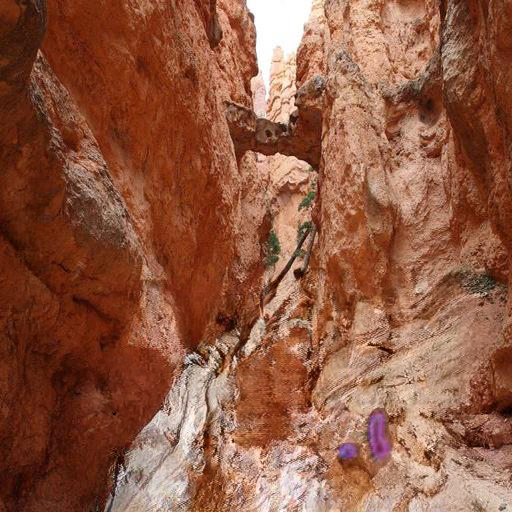}} &
\frame{\includegraphics[width=\MMM\linewidth]{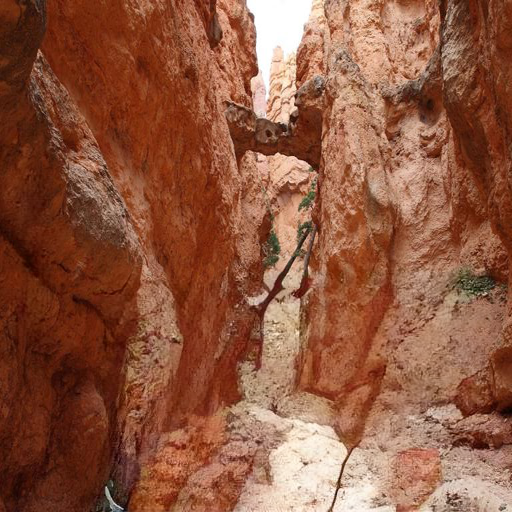}} &
\frame{\includegraphics[width=\MMM\linewidth]{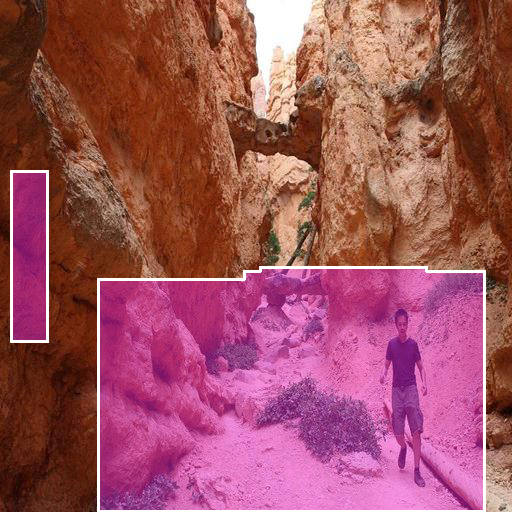}} \\

\frame{\includegraphics[width=\MMM\linewidth]{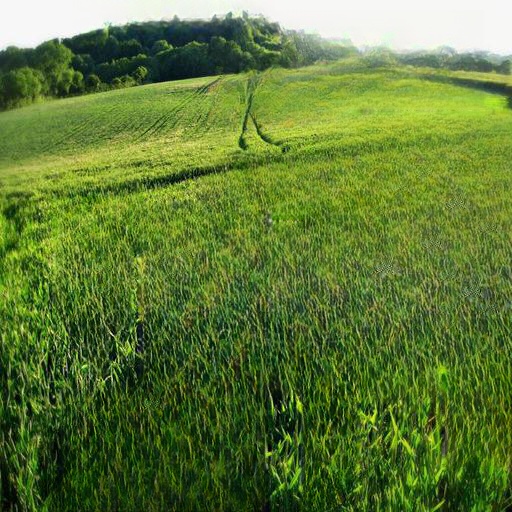}} &
\frame{\includegraphics[width=\MMM\linewidth]{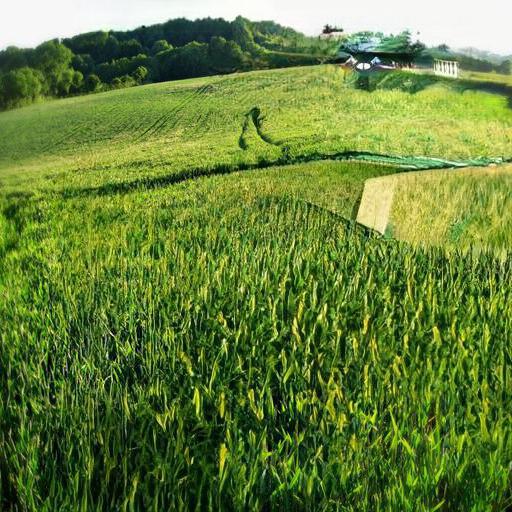}} &
\frame{\includegraphics[width=\MMM\linewidth]{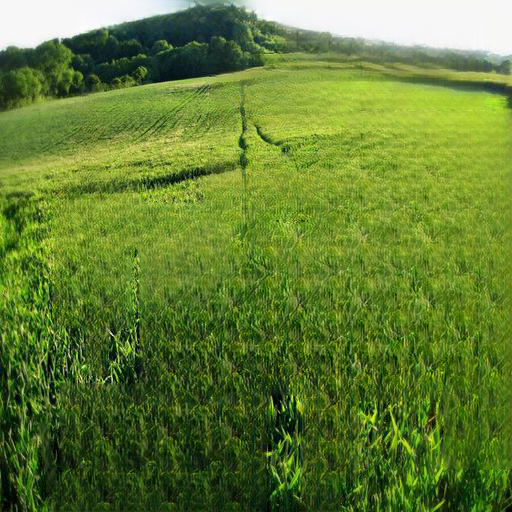}} &
\frame{\includegraphics[width=\MMM\linewidth]{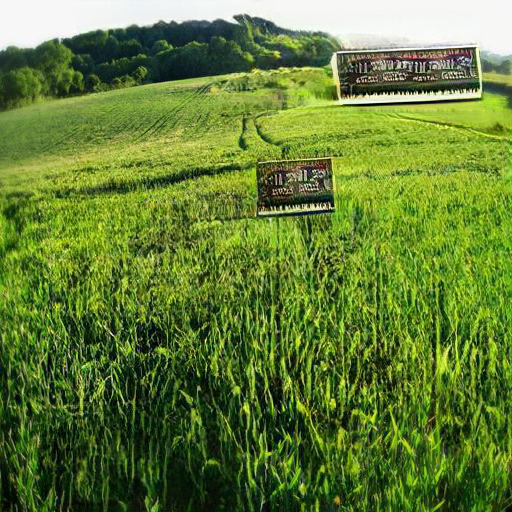}} &
\frame{\includegraphics[width=\MMM\linewidth]{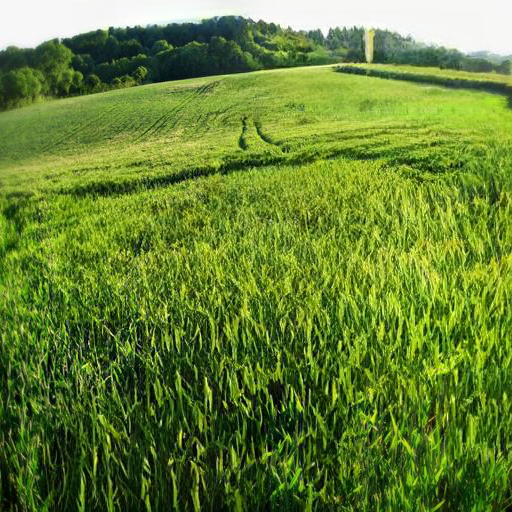}} &
\frame{\includegraphics[width=\MMM\linewidth]{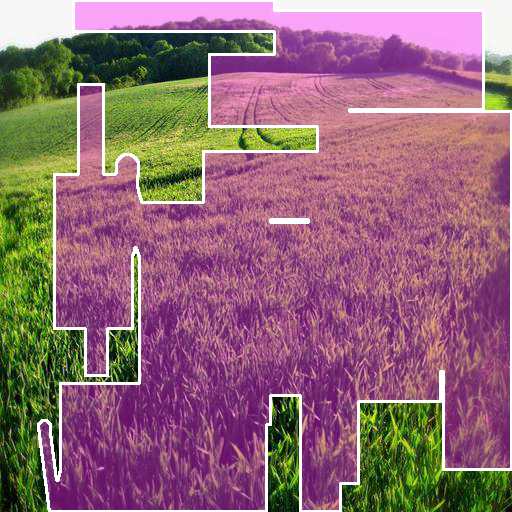}} \\

\frame{\includegraphics[width=\MMM\linewidth]{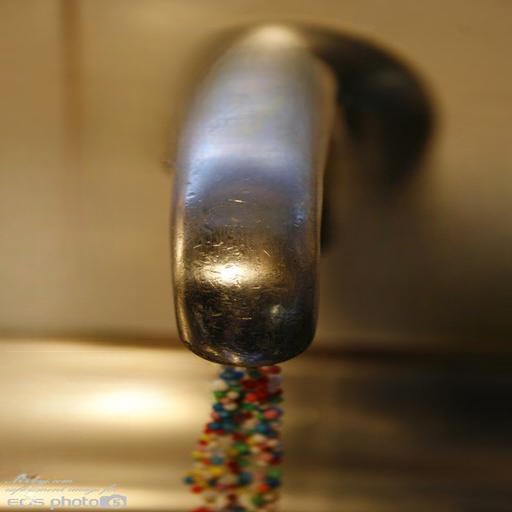}} &
\frame{\includegraphics[width=\MMM\linewidth]{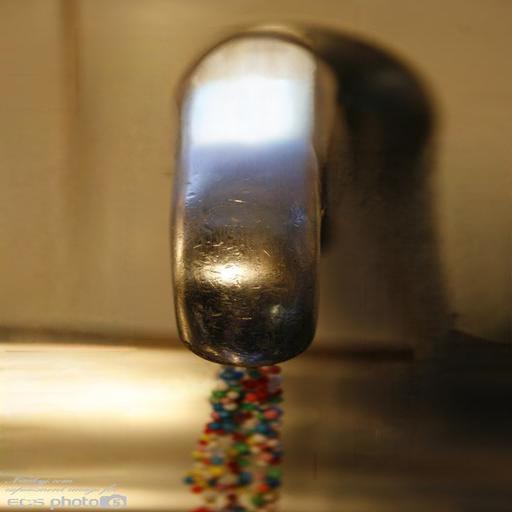}} &
\frame{\includegraphics[width=\MMM\linewidth]{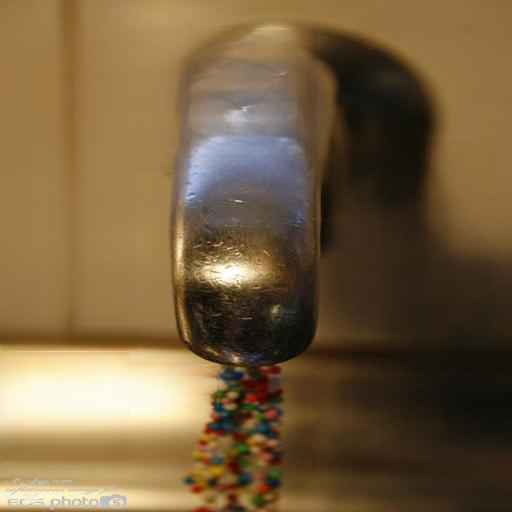}} &
\frame{\includegraphics[width=\MMM\linewidth]{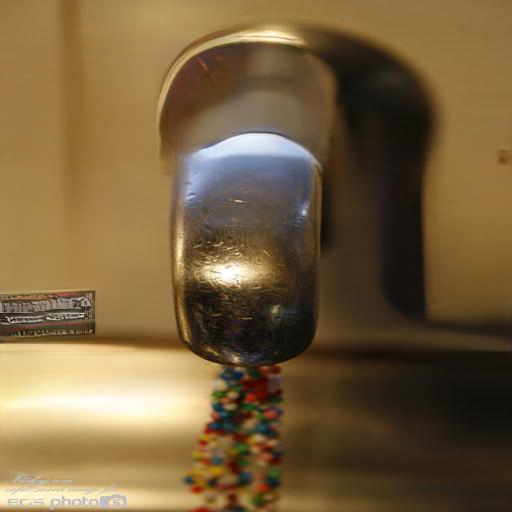}} &
\frame{\includegraphics[width=\MMM\linewidth]{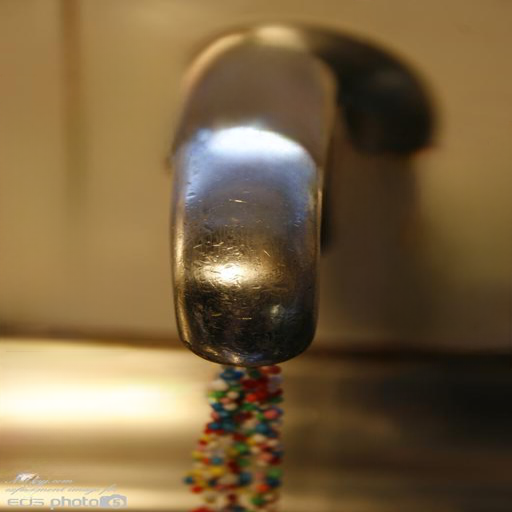}} &
\frame{\includegraphics[width=\MMM\linewidth]{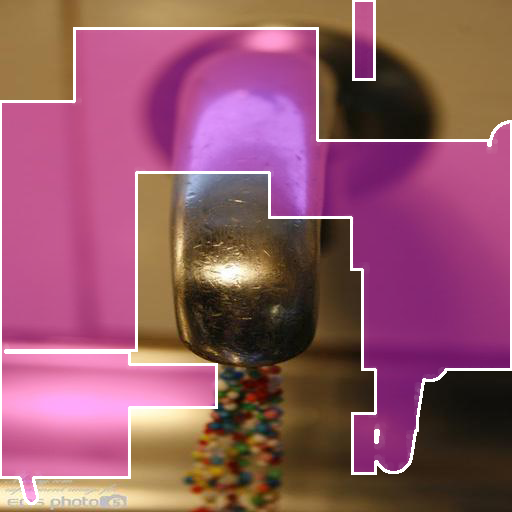}} \\

\frame{\includegraphics[width=\MMM\linewidth]{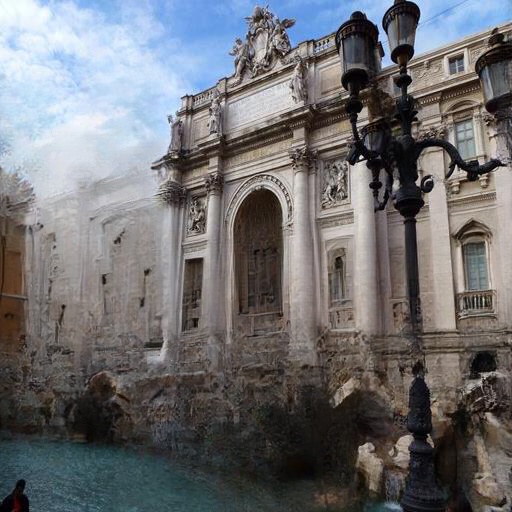}} &
\frame{\includegraphics[width=\MMM\linewidth]{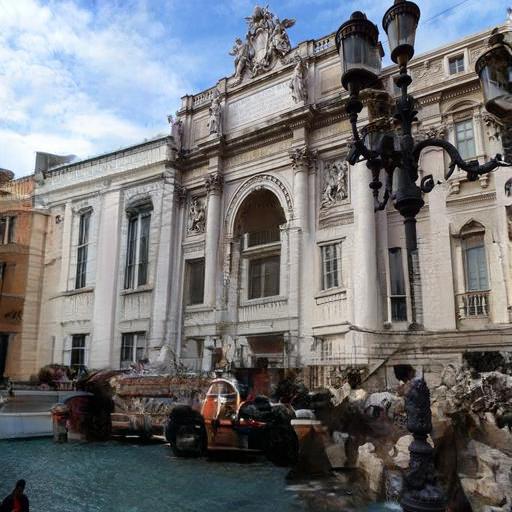}} &
\frame{\includegraphics[width=\MMM\linewidth]{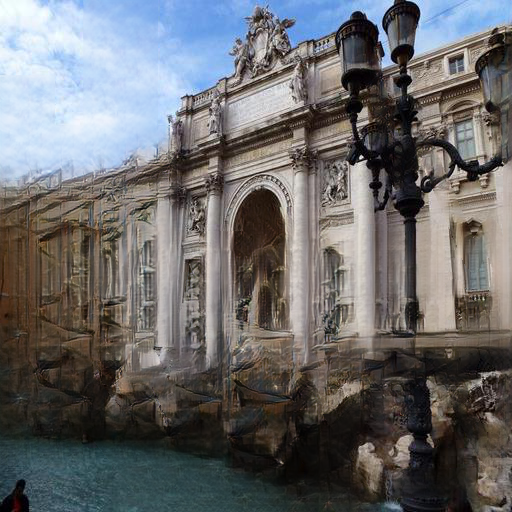}} &
\frame{\includegraphics[width=\MMM\linewidth]{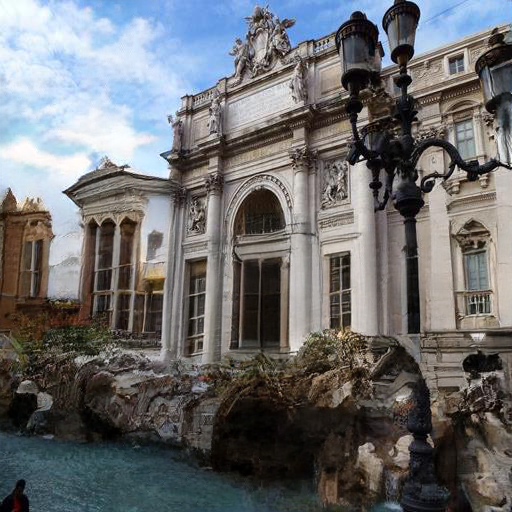}} &
\frame{\includegraphics[width=\MMM\linewidth]{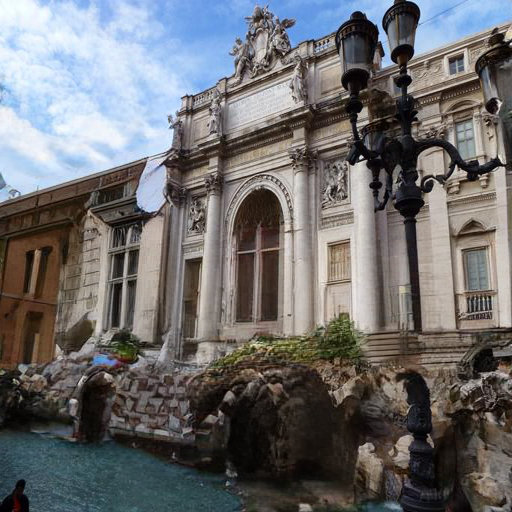}} &
\frame{\includegraphics[width=\MMM\linewidth]{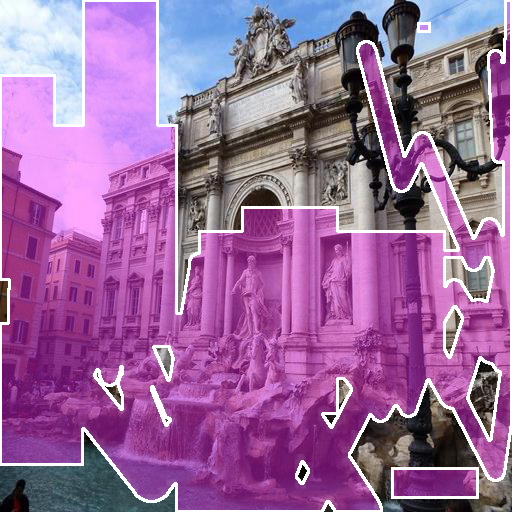}} \\

\end{tabular}
\caption{Qualitative comparison with the state-of-the-art image inpainting methods on Places2~\cite{places2} dataset. Zoom in for a better view.}
\label{fig:qualitative_image_places}
\end{figure*}

\end{document}